\newcommand{\ppname}{DDG-DA }
\newcommand{\ppnamenb}{DDG-DA}

\newcommand{\gfnamenb}{GF}
\newcommand{\eqwname}{RR }
\newcommand{\eqwnamenb}{RR}
\newcommand{\dataT}[1]{{\rm Data}^{(#1)}}
\newcommand{\amend}{}
\newcommand{\aamend}{}
\newcommand{\ignore}[1]{}

\def\hasappendix{1}
\def\extraExps{1}
\def\hasPseudocode{0}

\newcommand{\twoele}[2]{\begin{tabular}[l]{@{}l@{}}#1 \\ #2\end{tabular}} 

\def\year{2022}\relax
\documentclass[letterpaper]{article} 
\usepackage{aaai22}  
\usepackage{times}  
\usepackage{helvet}  
\usepackage{courier}  
\usepackage[hyphens]{url}  
\usepackage{graphicx} 
\usepackage{natbib}  
\usepackage{caption} 
\DeclareCaptionStyle{ruled}{labelfont=normalfont,labelsep=colon,strut=off} 
\usepackage{algorithm}
\usepackage{algorithmic}
\usepackage{multirow}
\usepackage{subfig}
\usepackage{amsfonts,bm}
\usepackage{dsfont}
\usepackage{color}
\usepackage{threeparttable}
\usepackage{amsmath}
\usepackage{amsthm}
\newtheorem{theorem}{Theorem}
\usepackage{booktabs}
\usepackage{placeins}  
\usepackage{tablefootnote}  
\usepackage{footnote}
\usepackage{newfloat}
\usepackage{listings}
\floatstyle{ruled}
\newfloat{listing}{tb}{lst}{}
\floatname{listing}{Listing}

\title{\ppnamenb: Data Distribution Generation for Predictable Concept Drift Adaptation}
\author {
    Wendi Li\textsuperscript{\rm 1,2}\equalcontrib, 
    Xiao Yang\textsuperscript{\rm 2}\equalcontrib,
    Weiqing Liu\textsuperscript{\rm 2},
    Yingce Xia\textsuperscript{\rm 2},
    Jiang Bian\textsuperscript{\rm 2}
}
\affiliations {
    \textsuperscript{\rm 1} University of Wisconsin–Madison\\
    \textsuperscript{\rm 2} Microsoft Research\\
    Wendi.Li@wisc.edu, \{Xiao.Yang, Weiqing.Liu, Yingce.Xia, Jiang.Bian\}@microsoft.com
}

\DeclareMathOperator*{\argmin}{arg\,min}
\begin{document}

\maketitle

\begin{abstract}
In many real-world scenarios, we often deal with streaming data that is sequentially collected over time. Due to the non-stationary nature of the environment, the streaming data distribution may change in unpredictable ways, which is known as concept drift. To handle concept drift, previous methods first detect when/where the concept drift happens and then adapt models to fit the distribution of the latest data. However, there are still many cases that some underlying factors of environment evolution are predictable, making it possible to model the future concept drift trend of the streaming data, while such cases are not fully explored in previous work.
In this paper, we propose a novel method \ppnamenb, that can effectively forecast the evolution of data distribution and improve the performance of models. Specifically, we first train a predictor to estimate the future data distribution, then leverage it to generate training samples, and finally train models on the generated data. We conduct experiments on three real-world tasks (forecasting on stock price trend, electricity load and solar irradiance) and obtain significant improvement on multiple widely-used models.

\end{abstract}

\section{Introduction}

\begin{figure}[t]
\centering
\includegraphics[width=\linewidth]{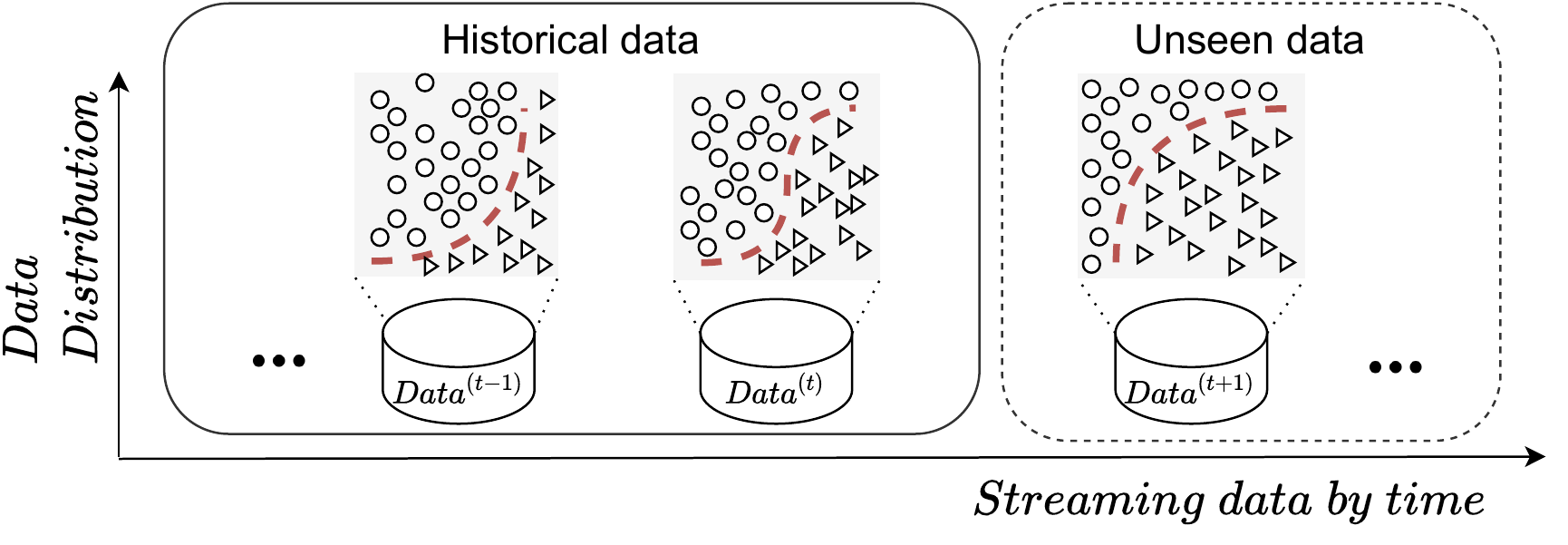}
\caption{An example of concept drifts on streaming data. Triangle and circle represent two classes of data. Data comes like a stream. The data distribution changes over time.}
\label{fig:drift}
\end{figure}

Machine learning algorithms have been widely applied to many real-world applications. One of the foundations of its success lies in assuming that the data is independent and identically distributed (briefly, the i.i.d. assumption). However, such an assumption is not always true for those learning tasks under broadly existing scenarios with streaming data. In a typical learning task over streaming data, as shown in Figure \ref{fig:drift}, the data comes in a sequential mode, meaning that, at any timestamp $t$, the learning task can only observe the new coming information, i.e., $\dataT{t}$, in addition to historical ones, i.e., $\dataT{t-1}$, $\dataT{t-2}$, $\cdots$, $\dataT{1}$. The goal of the task is usually to predict a certain target related to $\dataT{t+1}$. Due to the non-stationary nature of the real-world environment, the data distribution could keep changing with continuous data streaming. Such a phenomenon/problem is called \emph{concept drift} \cite{lu2018learning}, where the basic assumption is that concept drift happens unexpectedly and it is unpredictable \cite{gama2014survey} for streaming data.

To handle concept drift, previous studies usually leverage a two-step approach. Particularly, the first step is detecting the occurrence of concept drift in the streaming data, followed by the second step, if concept drift does occur, which adapts the model with new coming data by training a new model purely with latest data or making an ensemble between the new model with the current one, or fine-tuning the current model with latest data \cite{lu2018learning}. Most of these studies share the same assumption that the latest data contains more useful information w.r.t. the upcoming streaming data \cite{zhao2020handling}, and thus the detection of concept drift and following model adaptation is mainly applied to the latest data.

Therefore, existing methods for handling concept drift will detect concept drift on the latest arrived data $Data^{(t)}$ at timestamp $t$ and adapt the forecasting model accordingly.
The concept drift continues, and the adapted model on $Data^{(t)}$ will be used on unseen streaming data in the future (e.g., $Data^{(t+1)}$). The previous model adaptation has a one-step delay to the concept drift of upcoming streaming data, which means a new concept drift has occurred between timestamp $t$ and $t+1$.

\begin{figure}[htbp]
\centering
\subfloat[Stock price]{
\includegraphics[width=0.15\textwidth]{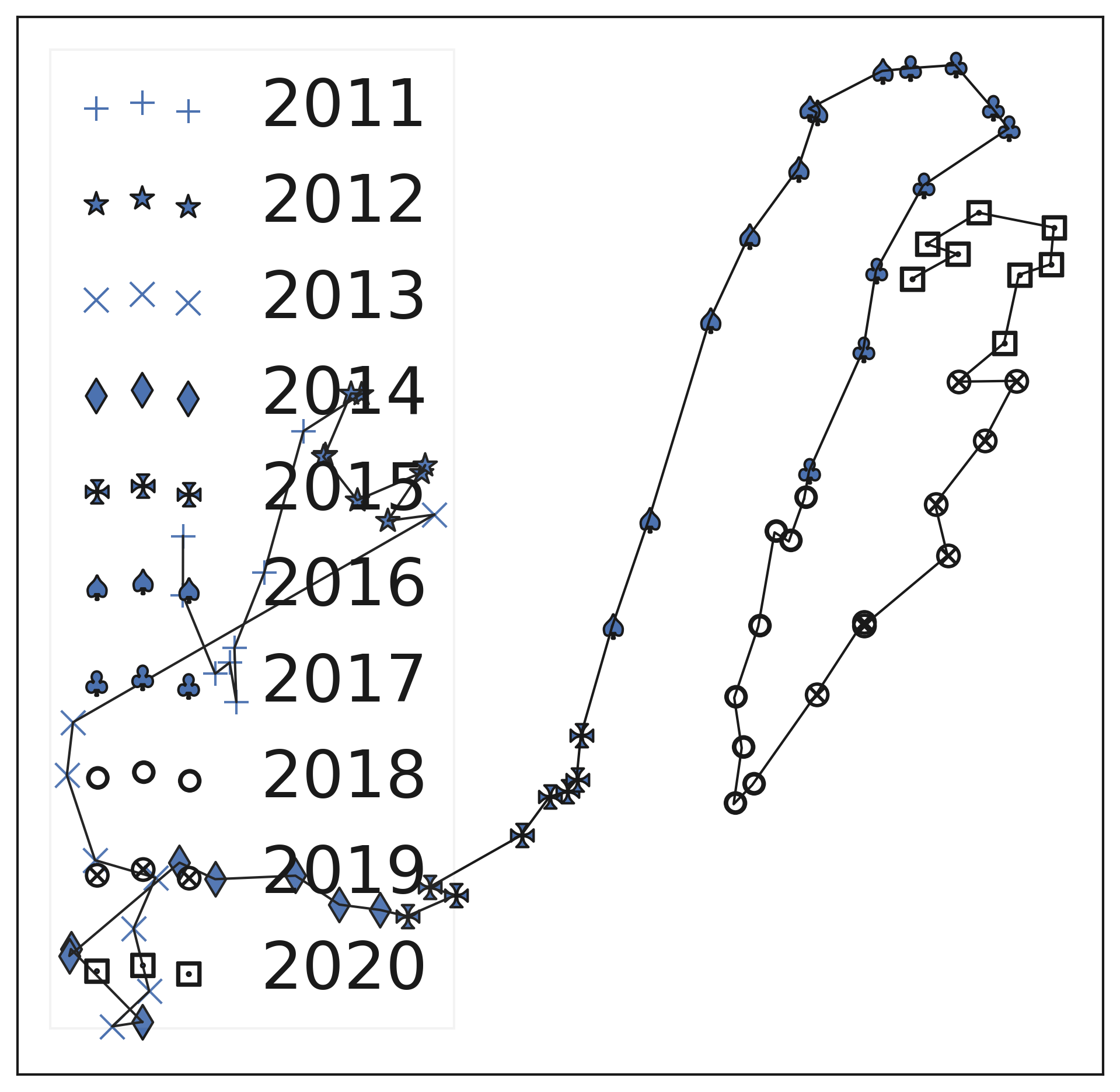}
}%
\subfloat[Electricity load]{
\includegraphics[width=0.15\textwidth]{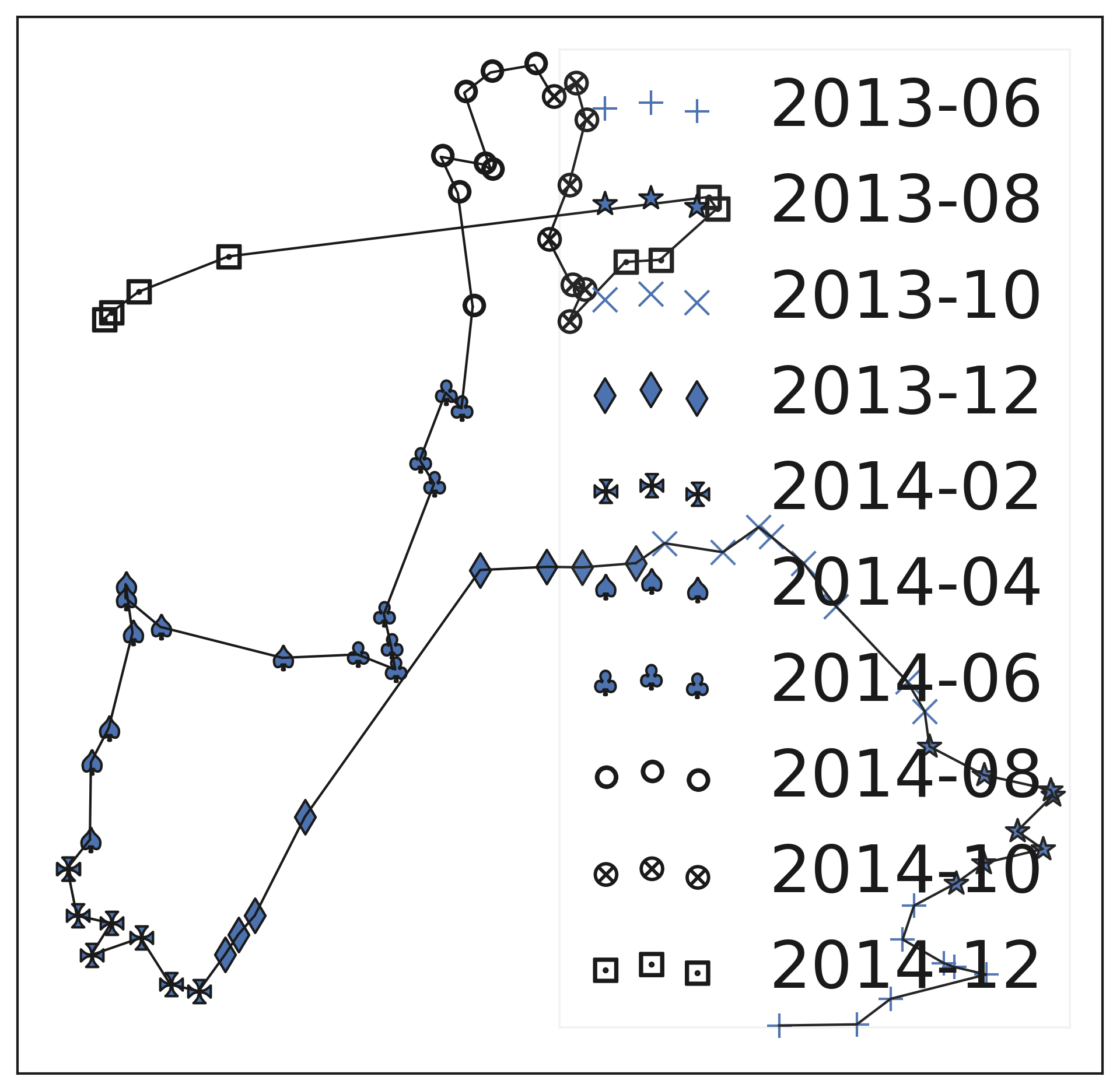}
}%
\subfloat[Solar irradiance]{
\includegraphics[width=0.15\textwidth]{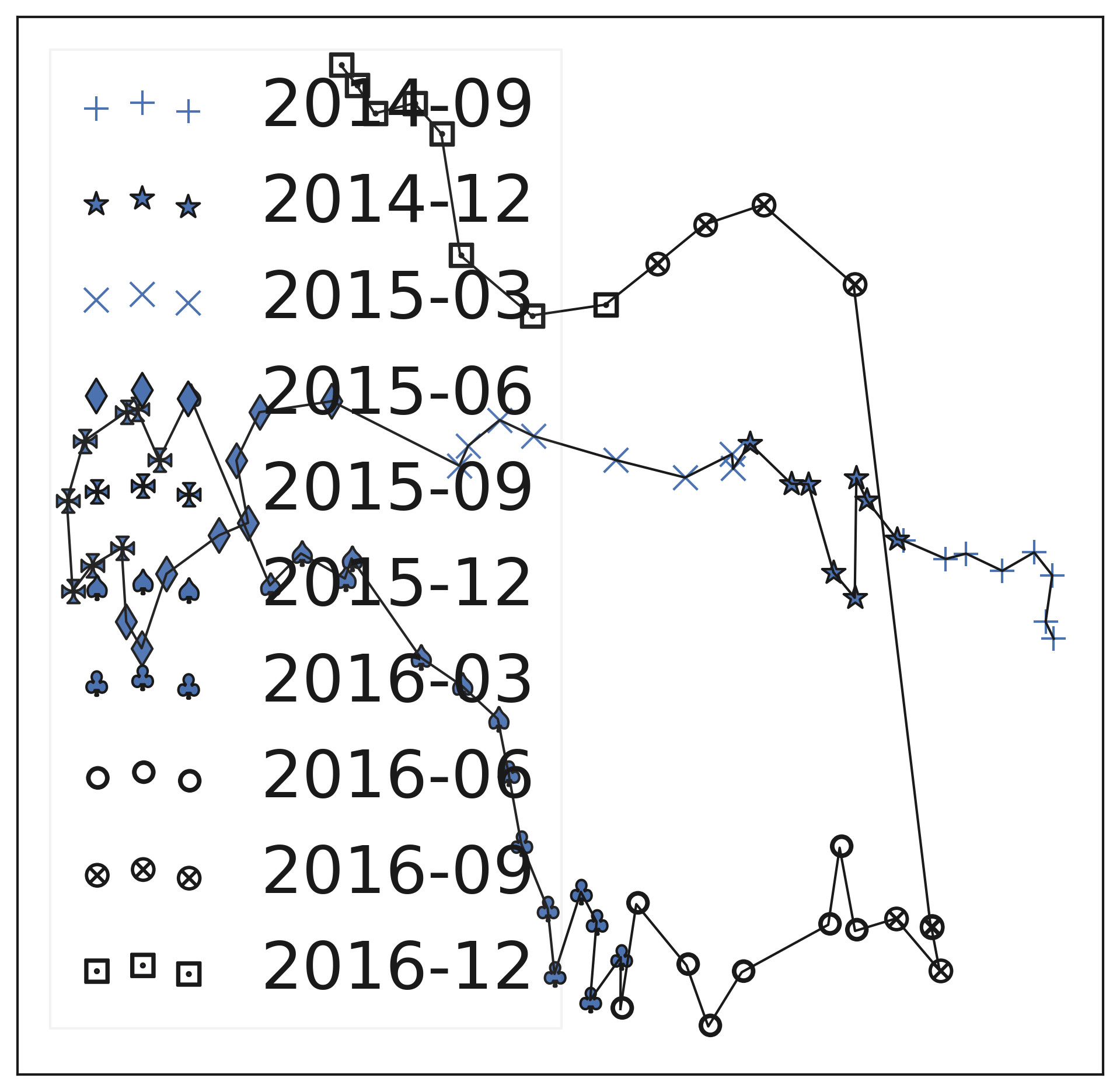}
}%
\caption{The visualization of concept drifts over time; though the concept drifts unexpectedly at some time, it drifts in a gradual nonrandom way most of the time. {\amend If the concept drifts are all sudden and do not follow any pattern, it may not be handled. However, in real-world scenarios, as we have shown, most concept drifts have a nonrandom trend rather than completely random. \ifodd\hasappendix Even for some quick concept drifts in the scenarios, the case study in Appendix \ref{Sec:FurtherExp} illustrates that \ppname can handle them to some extent. \fi}}
\label{fig:DataPattern}
\end{figure}

Nevertheless, apart from detecting unexpectedly occurred concept drift, most of the existing studies paid less attention to the scenarios that concept drift evolves in a gradual nonrandom way, which are in fact more common in streaming data.
In Figure \ref{fig:DataPattern}, we visualize the trace of concept drift in streaming data on three real-world tasks: forecasting on stock price trend, electricity load and solar irradiance \ifodd\hasappendix (detailed settings can be found in Appendix \ref{sec:dataDetails})\fi.  Each point indicates the data distribution (a.k.a. concept) represented by several features in a specific time period and is plotted on the figure with t-SNE \cite{van2008visualizing}. To better visualize the concept drift, we connect points contiguous in time to a line. 
From this figure, we can find that unexpected drastic concept drifts only occur occasionally. On the other hand, most of the points that are contiguous in time are spatially adjacent to each other and form a nonrandom trend, demonstrating a strong indication that a large portion of concept drifts do evolve in a gradual nonrandom way. More importantly, such gradual nonrandom concept drifts are in some sense predictable since patterns exist in the concept drift trend.

In this paper, we focus on predictable concept drift by forecasting the future data distribution.
We propose a novel method \ppnamenb\footnote{The code is available at \url{https://github.com/Microsoft/qlib/tree/main/examples/benchmarks_dynamic/DDG-DA}} to predict the data distribution of the next time-step sequentially, such that the model of the downstream learning task can be trained on the data sample from the predicted distribution instead of catching up with the latest concept drift only.
In practice, \ppname is designed as a dynamic data generator that can create sample data from previously observed data by following predicted future data distribution. In other words, \ppname generates the resampling probability of each historical data sample to construct the future data distribution in estimation.
However, it is quite challenging in reality to train this data generator to maximize the similarity between the predicted data distribution (represented by weighted resampling on historical data) and the ground truth future data distribution (represented by data in the future).
To address this challenge, we propose to first represent a data distribution by learning a model under this data distribution and then create a differentiable distribution distance to train the data generator. To verify the effectiveness of this approach, we also conduct a thorough theoretical analysis to prove the equivalence between traditional distribution distance, e.g. KL-divergence, and the proposed differentiable distribution distance \ifodd\hasappendix (Appendix \ref{Sec:theoretical})\fi. Furthermore, it is worth noting that our proposed new approach is quite efficient since the distribution distance is differentiable and could be easy for optimization.

Extensive experiments have been conducted on a variety of popular streaming data scenarios with multiple widely-used learning models to evaluate the effectiveness of our proposed
method \ppnamenb.
Experimental results have shown that \ppname could enhance the performance of learning tasks in all these scenarios.
Further analysis also reveals that \ppname can successfully predict the future data distribution in the circumstance of predictable concept drift. The predicted data distribution could benefit the learning task by proactively adapting to drifting concepts. \ifodd\hasappendix In Appendix \ref{Sec:casestudy}, more detailed case studies are presented to demonstrate that \ppname has learned reasonable and explainable knowledge. \fi
The code of \ppname is open-source on Github\footnote{\url{https://github.com/Microsoft/qlib}}.

To sum up, our contributions include
\begin{itemize}
    \item To the best of our knowledge, \ppname is the first method to model the evolving of data distribution in \emph{predictable concept drift} scenarios.
    \item We create a differentiable distribution distance and provide theoretical analysis to prove its equivalence to KL-divergence distribution distance.
	\item We conduct extensive experiments on different real-world concept-drift-predictable scenarios to show that \ppname outperforms SOTA concept drift adaptation methods by training models on resampled dataset(predicted data distribution).
\end{itemize}

\section{Background and Related Work}
\label{sec:back_related}

\subsection{Streaming Data and Concept Drift}
\subsubsection{General Concept Drift}
In a large number of practical scenarios, data are collected over time sequentially. The streaming data could be formalized as $\bm{X} = \{ \bm{x}^{(1)}, \bm{x}^{(2)}, \dots, \bm{x}^{(T)} \}$, where each element $\bm{x}^{(t)} \in  \mathds{R}^m$ pertaining to $\bm{X}$ is an array of $m$ values such that $\bm{x}^{(t)} = \{ x_{1}, x_{2}, \dots, x_{m} \}$. Each one of the $m$ scalar values corresponds to the input variable observed at time $t$.
Given a target sequence $\bm{y}=\{y^{(1)}, y^{(2)}, \dots, y^{(T)}\}$ corresponding to $\bm{X}$, algorithms are designed to build the model on historical data
$\{(\bm{x}^{(i)}, y^{(i)}) \}_{i=1}^{t}$ 
and forecast $\bm y$ on the future data $D^{(t)}_{test} = \{ (\bm{x}^{(i)}, y^{(i)}) \}_{i=t + 1}^{t + \tau}$ at each timestamp $t$.
Data are drawn from a joint distribution $(\bm{x}^{(t)}, y^{(t)}) \sim  p_t(\bm{x}, y)$. The goal of algorithms is to build a model to make accurate predictions on unseen data $D^{(t)}_{test}$ in the future.
Due to the non-stationary nature of the environment, the joint distribution $p_t$ is not static and changes over time, which is a well-known problem named \emph{concept drift}. Formally, the concept drift between two timestamps $t$ and $t + 1$ can be defined as $\exists \bm{x}: p_{t}(\bm{x}, y) \neq p_{t + 1}(\bm{x}, y)$.

To improve the forecasting accuracy in the future, adapting models to accommodate the evolving data distribution is necessary, which is the focused problem in this paper.
It can be formalized as
\begin{equation*}
\min_{f^{(t)}, f^{(t + 1)}, \dots, f^{(t + \tau)}} \sum_{i = t} ^ {t + \tau} l \left ( f^{(i)}(\bm{x}^{(i)}), y^{(i)}  \right )
\end{equation*}
where $f^{(t)}$ is learned from the training data in the previous stream
$ D^{(t)}_{train} = \{ (\bm{x}^{(i)}, y^{(i)}) \}_{i=t - k}^{t - 1}$ with window size $k$
and will be adapted continuously to minimize the target metric $loss$ in given time period $[t, t + \tau]$. 
In many practical scenarios, the streaming data come continually over time and might never end. Accommodating such an amount of data in memory will be infeasible. 
Therefore only a limited memory size $k$ is allowed for an online setting. Data in the memory could be used for model adaptation.

{\amend

\subsubsection{The Categorization of Concept Drifts and Our Scope}
The concept drifts can be categorized into different types when based on different criteria \cite{ren2018knowledge}. \cite{webb2016characterizing} provides quantitative measures to categorize concept drifts. For instance, a concept drift can be either \textit{abrupt} or \textit{gradual} according to its changing speed. It can also be \textit{recurring} or \textit{non-recurring} depending on whether similar concepts have appeared in the data stream.

Our method aims to handle scenarios with \textit{predictable} concept drifts defined in \cite{minku2009impact}, which refers to the concept drifts that follow a pattern rather than completely random due to the seasonality and reoccurring contexts. 
As \cite{ren2018knowledge, vzliobaite2016overview} state, the concept drifts in real-world evolving data are often caused by changes of the hidden contexts. Taking stock market, electricity load and solar irradiance as examples, because they are affected by some cyclic elements (time of the day, day of the week, month of the year, etc.), all of them have periodic, recurrent contexts. 
Abrupt drifts are rare due to nonrandom time-varying contexts, which is shown in Figure \ref{fig:DataPattern}. Such reasons give these concept drifts nonrandom trends rather than completely unpredictable ones.
\cite{khamassi2018discussion,webb2016characterizing} demonstrate that some characteristics of concept drift can be measured and predicted.
}

\subsection{Related Work}

Concept drift is a well-known topic in recent research works \cite{lu2018learning,gama2014survey}.
The concept drift research works usually focus on streaming data, which come in a streaming form and are usually collected in non-stationary environments as Figure \ref{fig:drift} shows.  {\amend In the setting of most of the previous works, concept drift happens unexpectedly and is unpredictable. Without  proactively predicting concept drift, these concept drift adaptation methods usually assume that the newer the data, the smaller the distribution gap between it and the future coming data. For instance, forgetting-based methods \cite{koychev2000gradual,klinkenberg2004learning} decay sample weights according to their age. When new labelled data are available, the model adaptation method detects concept drift and optionally updates the current model. However, the adapted new model will be used in the future, and the concept drift continues, which could make the model adaptation fail on unseen data in the future.

\subsubsection{Previous Methods to Handle Predictable Concept Drift}
Predictable concept drift is a scenario that the evolving concept drift is predictable to a certain extent \textbf{before concept drift happens}.  In this scenario, methods are encouraged to predict and adapt to the new concept before drift happens.
Proactively adapting models to the future joint distribution $p(\bm x, y)$ is the key to catching up with the concept evolution.
Though most previous works assume that concept drift happens unexpectedly and is unpredictable.
There are still a few of them noticing that concept drift is predictable \cite{minku2009impact,khamassi2018discussion} under certain additional assumptions.
Some case studies in real-world scenarios demonstrate that the prediction of concept drift is possible \cite{gama2014survey}.
\cite{yang2005combining} try to deal with recurrent concept drifts. They use a small trigger window to detect the concept drift. Based on the small amount of data in the window, they build a model to predict the new concept \textbf{after concept drift happens}. It is designed for the traditional concept drift setting instead of predictable concept drift.
\ppname tries to predict and adapt to the new concept \textbf{before it happens}.  As far as we know, we are the first method designed for predictable concept drift.

\subsubsection{Types of Methods to Handle Concept Drift}

According to the learning mode, methods can be categorized into \emph{retraining} \cite{bach2008paired} and \emph{incremental adaptation} \cite{alippi2008just}.
Retraining methods will discard the current model and learn a new model on a limited size of data.
Incremental adaptation methods will update the current model incrementally.

From another perspective, most adaptive learning methods are designed for specific models. In lots of practical forecasting tasks, customized models will be designed specifically. These model-specific adaptive learning methods can only be carried out on specific forecasting models and are not general solutions, which narrows their practicability in real applications.
Only a few adaptation methods are model-agnostic \cite{klinkenberg2004learning,koychev2002tracking,koychev2000gradual,vzliobaite2009combining}.  
\ppname focuses on predicting future data distribution,  which is a model-agnostic solution and can benefit different customized forecasting models on streaming data.
}

\section{Method Design}

\label{sec:method_design}

\begin{figure}[tb]
\centering
  \includegraphics[width=\linewidth]{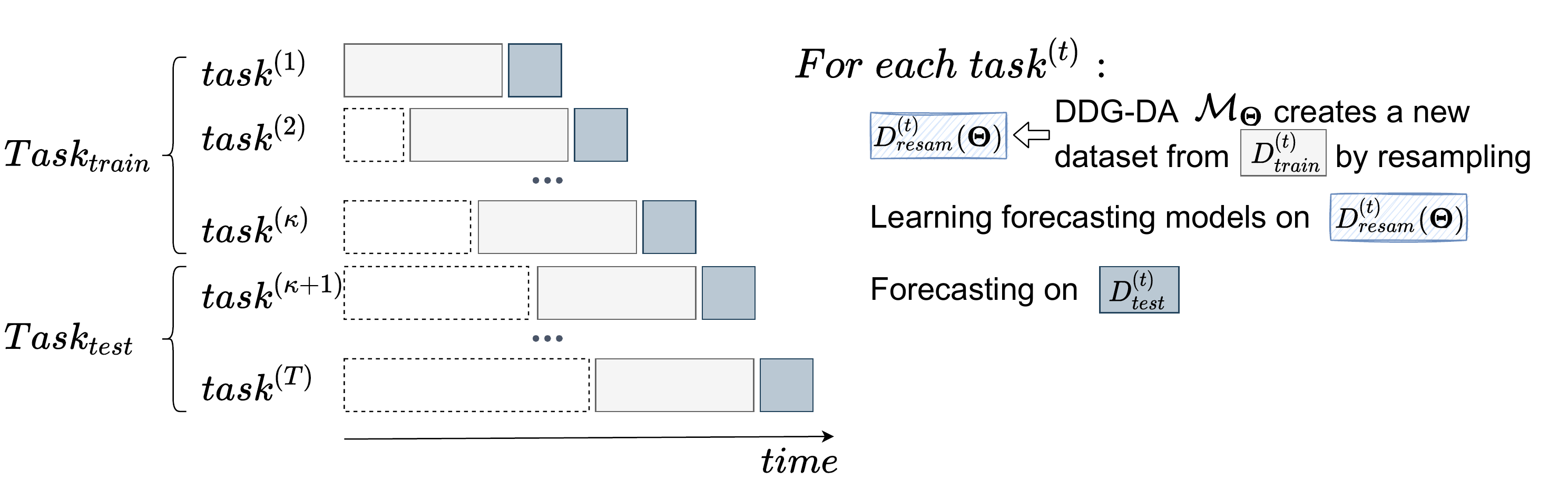}
  \caption{Training data (historical data) and test data (recent unseen data) change over time; the objective of each task is to improve the forecasting performance on test data. }
  \label{fig:task}
\end{figure}%

\ifodd\hasPseudocode
\begin{algorithm}[htb]
\caption{\ppname Learning Algorithm}
\label{alg:learn}
\begin{algorithmic}[1] 
\STATE \textbf{Input}: Historical streaming data $\mathcal{D}$
\STATE Create $Task_{train}$  from $\mathcal{D}$ via Fig. (\ref{fig:task})
\STATE Initialize parameters $\bm{\Theta}$ of \ppnamenb.
\WHILE{parameters $\bm{\Theta}$ are not converged }
\FOR{each $task^{(t)}$ in  $Task_{train}$}
\STATE $\{D^{(t)}_{train}, D^{(t)}_{test}\} \leftarrow task^{(t)}$
\STATE Resampling $D^{(t)}_{resam}(\bm{\Theta})$ from $D^{(t)}_{train}$ with prob. $q^{(t)}_{train} = \mathcal{M}_{\Theta} \left ( g(D^{(t)}_{train}) \right )$
\STATE Learn $y^{(t)}_{proxy}(\bm{x}; \phi^{(t)})$ as a proxy for $y^{(t)}_{resam} (\bm{x}; \bm{\Theta}^{(t)})$ to evaluate the distribution distance between $D^{(t)}_{resam}(\bm{\Theta})$ and $D^{(t)}_{test}$  via Eq.\eqref{eq:mse}
\STATE Add the distribution distance to the accumulated loss
\ENDFOR
\STATE Update $\bm{\Theta}$ via stochastic gradient descent
\ENDWHILE
\STATE \textbf{Return} \ppname $\mathcal{M}_{\bm{\Theta}}$
\end{algorithmic}
\end{algorithm}

\begin{algorithm}[htb]
\caption{\ppname Forecasting Algorithm}
\label{alg:infer}
\begin{algorithmic}[1] 
\STATE \textbf{Input}: $task^{(t)}\in Task_{test}$, \ppname $\mathcal{M}_{\bm{\Theta}}$
\STATE $\{D^{(t)}_{train}, D^{(t)}_{test}\} \leftarrow task^{(t)}$
\STATE Resample on $D^{(t)}_{train}$ to with prob. $q^{(t)}_{train} = \mathcal{M}_{\Theta} \left ( g(D^{(t)}_{train}) \right )$ to create  $D^{(t)}_{resam}(\bm{\Theta})$ 
\STATE Train forecasting model $f_{\bm{\theta}_{(t)}^*}$ on $D^{(t)}_{resam}(\bm{\Theta})$ 
\STATE $\hat{\bm{y}}^{(t)}_{test} \leftarrow f_{\bm{\theta}_{(t)}^*}(\bm{X}^{(t)}_{test})$    \COMMENT{ $\bm{X}^{(t)}_{test}$ is from $D^{(t)}_{test}$}
\STATE \textbf{Return} $\hat{\bm{y}}^{(t)}_{test}$
\end{algorithmic}
\end{algorithm}
\fi

\subsection{Overall Design}

\begin{figure*}[htbp]
\centering
  \includegraphics[width=\linewidth]{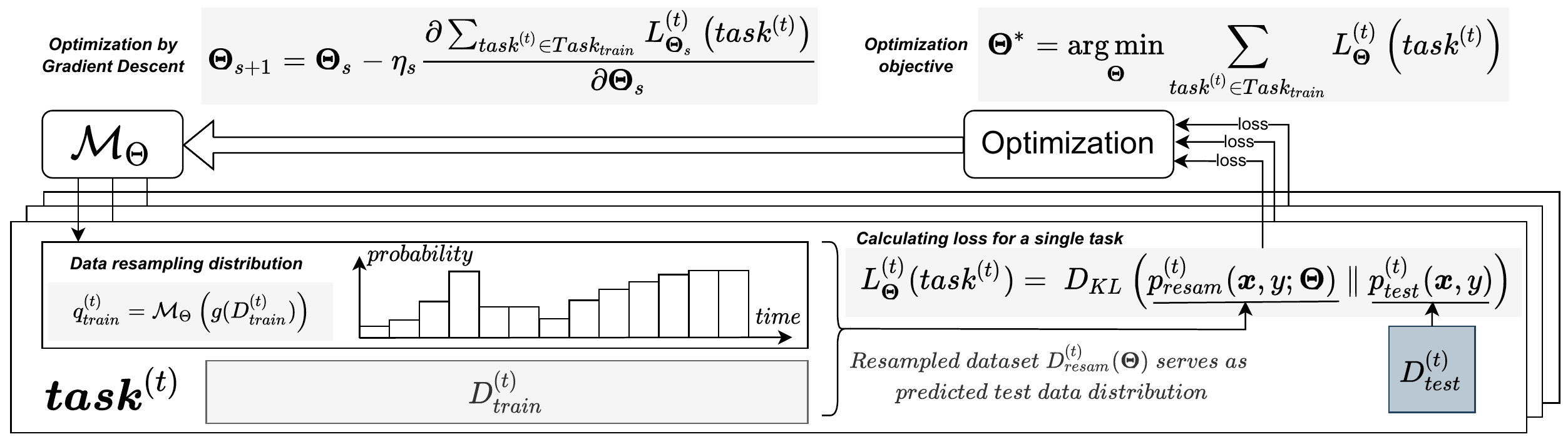}
  \caption{The learning process of \ppnamenb; \ppname $\mathcal{M}_{\Theta}$ learns to guide the training process of forecasting model by generating dataset $D^{(t)}_{resam}(\bm{\Theta})$ resampled from $D^{(t)}_{train}$ with probability $q_{train}^{(t)}$.  $q_{train}^{(t)}$ is the resampling probability given by $\mathcal{M}_{\Theta}$ at timestamp $t$.
  }
  \label{fig:framework}
\end{figure*}

In streaming data, forecasting models are trained and adapted on historical data (training data $D_{train}^{(t)}$) and make predictions on unseen data (test data $D_{test}^{(t)}$) in the future. 
As shown in Figure \ref{fig:task},  training data and test data change over time.
For each timestamp $t$,  the target of $task^{(t)} := ( D_{train}^{(t)}, D_{test}^{(t)} )$ is to learn a new model or adapt an existing model on historical data $D_{train}^{(t)}$ and minimize the loss on $D_{test}^{(t)}$. 
The data come continuously and might never end. Therefore models can leverage a limited size of $D_{train}^{(t)}$ at timestamp $t$ due to storage limitation.
$D_{train}^{(t)}$ is a dataset sampled from training data distribution $p_{train}^{(t)}(\bm{x}, y)$ and $D_{test}^{(t)}$ from test data distribution $p_{test}^{(t)}(\bm{x}, y)$.
The two distributions may be different (a.k.a. non-i.i.d.).
This distribution gap is harmful to the forecasting accuracy on $D_{test}^{(t)}$ when learning forecasting models on $D_{train}^{(t)}$ with a distribution different from $D_{test}^{(t)}$.

\subsubsection{\ppname Learning}
To bridge this gap, \ppname(annotated as $\mathcal{M}_{\Theta}$) tries to model the concept drift and predict the test data distribution $p^{(t)}_{test}(\bm{x}, y)$.
The framework of \ppname is demonstrated in Figure \ref{fig:task}.
\ppname will act like a weighted data sampler to resample on $D_{train}^{(t)}$ and create a new training dataset $D_{resam}^{(t)}(\bm{\Theta})$ whose data distribution is $p^{(t)}_{resam}(\bm{x}, y; \bm{\Theta})$ (the distribution of the resampled dataset serves as the prediction of test distribution). \ppname tries to minimize difference between the $p^{(t)}_{resam}(\bm{x}, y; \bm{\Theta})$ (the predicted data distribution) and the test data distribution $p^{(t)}_{test}(\bm{x}, y)$ (the ground truth data distribution).


During the training process of \ppnamenb, $\mathcal{M}_{\Theta}$ tries to learn patterns on $task^{(t)} \in Task_{train}$ by minimizing the data distribution distance between $p^{(t)}_{resam}(\bm{x}, y; \bm{\Theta})$ and $p^{(t)}_{test}(\bm{x}, y)$. The knowledge learned by $\mathcal{M}_{\Theta}$ from $Task_{train}$ is expected to transfer to new tasks from $Task_{test}$.

\subsubsection{\ppname Forecast}
For a given task $task^{(t)} \in Task_{test}$,  
the forecasting model is trained on dataset $D_{resam}^{(t)}(\bm{\Theta})$  under distribution $p^{(t)}_{resam}(\bm{x}, y; \bm{\Theta})$ and forecasts on test data $D_{test}^{(t)}$.
$p^{(t)}_{resam}(\bm{x}, y; \bm{\Theta})$ is the distribution of resampled dataset $D_{resam}^{(t)}(\bm{\Theta})$ and the decisive factor to the trained forecasting model
. With the help of \ppnamenb, the distribution of $D_{resam}^{(t)}(\bm{\Theta})$ is more similar to $D_{test}^{(t)}$ than $D_{train}^{(t)}$.
The forecasting model's performance  
on $D_{test}^{(t)}$ is expected to be improved.

\subsubsection{A Practical Example}
To make our method more understandable, we explain it with an example in the stock price trending forecasting scenario.
To handle the concept drift in data, we retrain a new model each month (the rolling time interval is 1 month) based on two years of historical data (memory size is limited in an online setting). Each chance to retrain a new model to handle concept drift is called a \emph{task}. For example, the first $task^{(2011/01)}$ contains training  data $D^{(2011/01)}_{train}$  from 2009/01 to 2010/12 and a month of test data $D^{(2011/01)}_{test}$ in 2011/01. \ppname creates $D^{(2011/01)}_{resam}(\bm{\Theta})$ based on $D^{(2011/01)}_{train}$ to train a forecasting model to achieve better performance on $D^{(2011/01)}_{test}$.
A new task is created in each month.
These tasks are arranged in chronological order and separated at the beginning of 2016 into $Task_{train}$ (all $D^{(t)}_{test}$ in $Task_{train}$ range from 2011 to 2015) and $Task_{test}$ (all $D^{(t)}_{test}$ in $Task_{test}$ range from 2016 to 2020). 
\ppname is trained on $Task_{train}$ (learning to generate $D^{(t)}_{resam}(\bm{\Theta})$ and minimize its distribution distance with $D^{(t)}_{test}$). Then \ppname is evaluated on $Task_{test}$.

\subsection{Model Design and Learning Process}

The overall model design and learning process of \ppname are shown in Figure \ref{fig:framework}. \ppname will build a set of tasks $Task_{train} = \{ task^{(1)}, \dots, task^{(\tau)} \}$ for training \ppnamenb. The goal of the learned \ppname is to improve the performance on test tasks $Task_{test} := \{ task^{(\tau + 1)}, \dots, task^{(T)} \}$.

\subsubsection{Feature Design}
\label{sec:feature_design}

\ppname is expected to guide the model learning process in each $task^{(t)}$ by forecasting test data distribution. Historical data distribution information is useful to predict the target distribution of $D_{test}^{(t)}$ and is input into \ppnamenb. \ppname will learn concept drift patterns from training tasks and help to adapt models in test tasks. 

\ppname could be formalized as $q^{(t)}_{train} = \mathcal{M}_{\Theta} \left ( g(D^{(t)}_{train}) \right )$.  $g$ is a feature extractor. It takes $D^{(t)}_{train}$ as input and outputs historical data distribution information. $\mathcal{M}_{\Theta}$ leverages the extracted information and output the resampling probabilities for samples in $D^{(t)}_{train}$.
\ifodd\hasappendix More details about the feature design are attached in Appendix \ref{sec:ModelDetails}. \fi

\subsubsection{Objective Function}
$\mathcal{M}_{\Theta}$ accepts the extracted feature and outputs the resampling probability on $D^{(t)}_{train}$.
 The resampled dataset's joint distribution $p^{(t)}_{resam}(\bm{x}, y; \bm{\Theta})$ serves as the distribution prediction. 
The learning target of \ppname is to minimize the difference between $p^{(t)}_{resam}(\bm{x}, y; \bm{\Theta})$ and $p^{(t)}_{test}(\bm{x}, y)$. 
We focus on the most important concept drift subject $p(y\mid\bm{x})$ \cite{kelly1999impact} and assume the difference between $p^{(t)}_{test}(\bm{x})$ and $p^{(t)}_{resam}(\bm{x}; \bm{\Theta})$ are minor.
The loss of \ppname could be reformulated as
{\amend
\begin{align} \label{eq:loss_exp}
\small
\begin{split}
    & L_{\bm{\Theta}}(task^{(t)}) \\
    =& \; D_{KL} \left( p^{(t)}_{test}(\bm{x}, y)   \parallel p^{(t)}_{resam}(\bm{x}, y; \bm{\Theta} )  \right) \\
    =&  \; \mathbb{E}_{\bm{x} \sim  p^{(t)}_{test}(\bm{x})} 
    \left [ D_{KL} \left(p^{(t)}_{test}(y \mid  \bm{x} ) \parallel p^{(t)}_{resam}(y \mid \bm{x}  ; \bm{\Theta} )\right)\right ]  
\end{split}
\normalsize
\end{align}
}
where $D_{KL}$ represents the Kullback–Leibler divergence.

Normal distribution assumption is reasonable for unknown variables and often used in maximum likelihood estimation \cite{goodfellow2016deep}. Under this assumption,
$p^{(t)}_{test}(y \mid  \bm{x} ) = \mathcal{N}(y^{(t)}_{test}(\bm{x}), \sigma) $
and
$p^{(t)}_{resam}(y \mid  \bm{x} ) = \mathcal{N}(y^{(t)}_{resam}(\bm{x}; \bm{\Theta}), \sigma)$ where $\sigma$ is a constant.

The estimated expectation of Equation \eqref{eq:loss_exp} on empirical sampled dataset can be reformulated \ifodd\hasappendix (Theorem \ref{theo:01} in Appendix \ref{Sec:theoretical}) \fi as
\small
\begin{equation} \label{eq:mse}
    L_{\bm{\Theta}}(task^{(t)}) = \frac{1}{2} \sum_{(\bm{x}, y) \in D^{(t)}_{test} 
    }  
       \left \| y^{(t)}_{resam}(\bm{x} ; \bm{\Theta}) - y \right \|^2
\end{equation}
\normalsize

Summarizing losses of all the training tasks, the optimization target of \ppname could be formalized as 
\begin{equation} \label{eq:metamodel}
    \bm{\Theta} ^{*} = \mathop{\arg\min}_{\bm{\Theta}} \sum_{ task^{(t)} \in Task_{train}} L_{\bm{\Theta}} \left ( task^{(t)} \right )
\end{equation}

\ppname learns knowledge from $Task_{train}$ and transfers it to unseen test tasks $Task_{test}$.
In each task, \ppname forecasts the future data distribution and generate dataset $D^{(t)}_{resam}(\bm{\Theta})$. Learning the forecasting models on $D^{(t)}_{resam}(\bm{\Theta})$, it adapts to upcoming streaming data better.

\subsubsection{Optimization}

In this section, we will introduce the optimization process of \eqref{eq:mse}.
$y^{(t)}_{test}(\bm{x})$ can be get directly from dataset $D^{(t)}_{test}$.
To approximate $y^{(t)}_{resam}(\bm{x}; \bm{\Theta})$, \ppname builds a regression proxy model $y_{proxy}(\bm{x}; \phi^{(t)})$ on $D^{(t)}_{resam}(\bm{\Theta})$. 
The optimization of the proxy model can be formulated as 
\small
\begin{equation} \label{eq:subject}
    \bm{\phi}^{(t)} = \mathop{\arg\min}_{\bm{\phi}} \sum_{(\bm{x}, y) \in D^{(t)}_{resam}(\bm{\Theta})}  \left \| y_{proxy}(\bm{x}; \bm{\phi}) - y \right \|^2
\end{equation}
\normalsize

The learning process of \ppname becomes a \emph{bi-level optimization problem} \cite{gould2016differentiating}. The goal of the upper-level is Equation \eqref{eq:mse} and \eqref{eq:metamodel} ($y^{(t)}_{resam} (\bm{x}; \bm{\Theta})$ is replaced by $y_{proxy}(\bm{x}; \bm{\phi}^{(t)})$). The goal of the lower-level optimization Equation \eqref{eq:subject} can be regarded as a constraint. 

\small
{\aamend The overall bi-level optimization formulation of the \ppname is
\begin{equation}
\begin{aligned}
\argmin_{\bm{\Theta}}\quad & \sum_{task^{(t)}\in Task_{train}}\left(\sum_{(\bm{x},y)\in D_{test}^{(t)}}\|y_{proxy}(\bm{x};\bm{\phi}^{(t)})-y\|^2\right)\\
\textrm{s.t.}\quad & \bm{\phi}^{(t)} =\argmin_{\bm{\phi}}\sum_{(\bm{x}',y')\in D_{resam}^{(t)}(\bm{\Theta})}\|y_{proxy}(\bm{x}'; \bm{\phi})-y'\|^2\\
\end{aligned}
\end{equation}
where $\bm{\Theta}$ is the parameters of \ppnamenb.
}
\normalsize

Many methods have been proposed to solve such a bi-level optimization problem \cite{gould2016differentiating}.
Some methods based on hyperparameter optimization \cite{bengio2000gradient,baydin2014automatic,maclaurin2015gradient} are proposed. But they have limitations on either the network size or optimization accuracy.

One of the key barriers that stop researchers from optimizing Equation \eqref{eq:mse} directly is that the lower-level optimization Equation \eqref{eq:subject} usually can not be solved in a closed-form. The $\argmin$ operator is usually not differentiable, which makes some popular and efficient optimization algorithms (such as gradient-descend-based methods) impossible in the optimization of the upper-level (Equation \eqref{eq:metamodel}).
\cite{lee2019meta,bertinetto2018meta} argue that \cite{bengio2000gradient,baydin2014automatic} it is too costly and adopt algorithms with closed-form solutions in the lower-level optimization.

\begin{table*}[htbp]
\centering
\begin{threeparttable}
\begin{tabular}{l|rrrrr|rr|rrr}
\toprule
\multirow{2}{*}{Method} & \multicolumn{5}{c|}{Stock Price Trend Forecasting} & \multicolumn{2}{c|}{Electricity Load} & \multicolumn{3}{c}{Solar Irradiance}\\
  & \textit{IC} & \textit{ICIR} & \textit{Ann.Ret.} & \textit{Sharpe} & \textit{MDD} & \textit{NMAE} & \textit{NRMSE} & \textit{Skill} (\%) & \textit{MAE} & \textit{RMSE}\\
\midrule
\eqwname & 0.1178 & 1.0658 & 0.1749 & 1.5105 & -0.2907 & 0.1877 & 0.9265 & 7.3047 & 21.7704 & 48.0117\\
\gfnamenb-Lin & 0.1227 & \underline{1.0804} & 0.1739 & 1.4590 & -0.2690 & 0.1843 & 0.9109 & \underline{9.3503} & 21.6878 & \underline{46.9522}\\
\gfnamenb-Exp & 0.1234 & 1.0613 & 0.1854 & 1.5906 & -0.2984 & 0.1839 & 0.9084 & 9.2652 & 21.6841 & 46.9963\\
ARF & 0.1240 & 1.0657 & 0.1994 & 1.8844 & \textbf{-0.1176} & \underline{0.1733} & \underline{0.8901} & 8.6267 & \underline{21.0962} & 47.3270\\
Condor & \underline{0.1273} & 1.0635 & \underline{0.2157} & \underline{2.1105} & -0.1624 & --------- & --------- & --------- & --------- & ---------\\
\ppnamenb & \textbf{0.1312} & \textbf{1.1299} & \textbf{0.2565} & \textbf{2.4063} & \underline{-0.1381} & \textbf{0.1622} & \textbf{0.8498} & \textbf{12.1327}  & \textbf{18.7997} & \textbf{45.5110}\\

\bottomrule
\end{tabular}
\caption{Performance comparison of the concept drifts adaptation methods.  }
\label{tab:compareBaselines}
\end{threeparttable}
\end{table*}

\ppname adopts a model with a closed-form solution as $y^{(t)}_{proxy}(\bm{x}; \phi^{(t)})$. We have many choices, such as logistic regression \cite{kleinbaum2002logistic}, kernel-based non-linear model \cite{liu2003kernel} and differentiable closed-form solvers \cite{bertinetto2018meta}.
We choose a linear model $h(\bm{x};\bm{\phi}_{(t)}) = \bm{x} \bm{\phi}^{(t)} $ for $y_{proxy}(\bm{x}; \phi^{(t)})$ for simplicity. 
The resampling probability $q^{(t)}_{train}$ outputted by $\mathcal{M}_{\Theta}$ could be regarded as sample weights when learning forecasting models. 
The loss function in Equation \eqref{eq:subject} could be formulated as 
\small
\begin{equation*}
\begin{aligned}
l_{\bm{\phi}^{(t)}}\left(D^{(t)}_{resam}(\bm{\Theta})\right) =& \frac{1}{2} \sum_{(\bm{x}, y) \in D^{(t)}_{train}} q^{(t)}_{train} \left(\bm{x} \bm{\phi}^{(t)} - y  \right)^2 \\
=&  \frac{1}{2} \left(\bm{X}^{(t)} \bm{\phi}^{(t)} - \bm{y}^{(t)}\right)^{\top} \bm{Q}^{(t)} (\bm{X}^{(t)} \bm{\phi}^{(t)} - \bm{y}^{(t)})  \\
\end{aligned}
\end{equation*}
\normalsize
where $\bm{X}^{(t)}$, $\bm{y}^{(t)}$ and $\bm{Q}^{(t)}$ represent the concatenated features, labels and resampling probability in $D^{(t)}_{train}$.

It equals to a \textit{weighted linear regression} problem \cite{ruppert1994multivariate}. $\bm{\phi}^{(t)}$ has a closed-form solution formalized as
\begin{align} \label{eq:proxymodel}
\begin{split}
\bm{\phi}_{(t)}=& \; \mathop{\arg\min}_{\bm{\phi}}  \; l_{\bm{\phi}}(D^{(t)}_{resam}(\bm{\Theta})) \\
=& \; \left(   \left(\bm{X}^{(t)}\right)^\top  \bm{Q}^{(t)} \bm{X}^{(t)}\right)^{-1} \left(\bm{X}^{(t)}\right)^\top  \bm{Q}^{(t)} \bm{y}^{(t)}
\end{split}
\end{align}
\normalsize

This closed-form solution to Equation \eqref{eq:subject} makes the distribution distance differentiable. It makes the optimization objective of $\mathcal{M}_{\Theta}$ (Equation \eqref{eq:metamodel}) differentiable. Therefore, simple and efficient optimization algorithms can be used to train \ppname (e.g. stochastic gradient descent).


\section{Experiments}
\label{sec:exp}


In this section, we conduct experiments aiming to answer the following  research questions:
\begin{itemize}
    \item \textbf{{Q1}}: Can \ppname outperform SOTA concept drift adaptation methods in predictable concept drift scenarios?
    \item \textbf{{Q2}}: Can \ppname generalize to different forecasting models in different scenarios?
\end{itemize}

\ifodd\hasappendix
Further experiments are conducted in Appendix \ref{Sec:FurtherExp}. (1) Execution time of different methods; (2) Experiments in a concept-drift-\textbf{unpredictable} scenario to show our limitation; (3) Case studies of \ppnamenb. \fi


\subsection{Experiment Setup}\label{sec:exp_settings}
The experiments are conducted on multiple datasets in three real-world popular scenarios (forecasting on stock price trend, electricity load and solar irradiance  \cite{grinold2000active,pedro2019comprehensive}).
\ifodd\hasappendix The detailed experiment settings are described in Appendix \ref{sec:dataDetails}. \fi

\begin{table*}
\centering
\begin{threeparttable}
\resizebox{0.98\textwidth}{!}{%
\centering
\begin{tabular}{ll|rrrrr|rr|rrr}
\toprule
\multirow{2}{*}{Model} & \multirow{2}{*}{Method} & \multicolumn{5}{c|}{Stock Price Trend Forecasting} & \multicolumn{2}{c|}{Electricity Load} & \multicolumn{3}{c}{Solar Irradiance}\\
 &  & \textit{IC} & \textit{ICIR} & \textit{Ann.Ret.} & \textit{Sharpe} & \textit{MDD} & \textit{NMAE} & \textit{NRMSE} & \textit{Skill} (\%) & \textit{MAE} & \textit{RMSE}\\
\midrule
\textbf{Linear}
& \eqwname & 0.0859 & 0.7946 & 0.1578 & 1.4211 & \underline{-0.1721} & 0.2080 & 1.0207 & \underline{1.4133} & \underline{24.0910} & \underline{51.0632} \\
& \gfnamenb-Exp & \underline{0.0863} & \underline{0.8018} & \underline{0.1632} & \underline{1.4373} & \textbf{-0.1462} & \underline{0.2009} & \underline{0.9787} & 1.3660 & 24.1391 & 51.0877\\
& \ppnamenb & \textbf{0.0971} & \textbf{0.9193} & \textbf{0.1763} & \textbf{1.5733} & -0.2130 & \textbf{0.1973} & \textbf{0.9702} & \textbf{3.7378} & \textbf{23.6901} & \textbf{49.8592}\\
\midrule
\textbf{MLP}
& \eqwname & \underline{0.1092} & 0.8647 & 0.1803 & 1.5200 & \underline{-0.1797} & 0.1928 & 0.9682 & 8.2145 & 22.1285 & 47.5405\\
& \gfnamenb-Exp & 0.1091 & \underline{0.8654} & \underline{0.1889} & \underline{1.6121} & \textbf{-0.1738} & \underline{0.1898} & \underline{0.9588} & \underline{8.4668} & \underline{21.6236} & \underline{47.4098} \\
& \ppnamenb & \textbf{0.1211} & \textbf{0.9921} & \textbf{0.2181} & \textbf{1.9409} & -0.1864 & \textbf{0.1882} & \textbf{0.9537} & \textbf{10.0341} & \textbf{20.3422} & \textbf{46.5980}\\
\midrule
\multirow{2}{*}{{\twoele{\textbf{Light-}}{\textbf{GBM}}}}
& \eqwname & 0.1178 & \underline{1.0658} & 0.1749 & 1.5105 & \underline{-0.2907} & 0.1877 & 0.9265 & 7.3047 & 21.7704 & 48.0117\\
& \gfnamenb-Exp & \underline{0.1234} & 1.0613 & \underline{0.1854} & \underline{1.5906} & -0.2984 & \underline{0.1839} & \underline{0.9084} & \underline{9.2652} & \underline{21.6841} & \underline{46.9963}\\
& \ppnamenb & \textbf{0.1312} & \textbf{1.1299} & \textbf{0.2565} & \textbf{2.4063} & \textbf{-0.1381} & \textbf{0.1622} & \textbf{0.8498} & \textbf{12.1327} & \textbf{18.7997} & \textbf{45.5110}\\
\midrule
\ifodd \extraExps
\multirow{2}{*}{{\twoele{\textbf{Cat-}}{\textbf{Boost}}}}
& \eqwname & 0.1047 & 0.7088 & 0.1782 & 1.3246 & -0.2860 & 0.1899 & 0.9304 & 9.9895 & 20.5114 & 46.6211\\
& \gfnamenb-Exp & \underline{0.1058} & \underline{0.7114} & \underline{0.1879} & \underline{1.3968} & \underline{-0.2414} & \underline{0.1884} & \underline{0.9291} & \underline{10.0384} & \underline{20.4882} & \underline{46.5958}\\
& \ppnamenb & \textbf{0.1173} & \textbf{0.8386} & \textbf{0.1908} & \textbf{1.5787} & \textbf{-0.2102} & \textbf{0.1811} & \textbf{0.8905} & \textbf{10.4187} & \textbf{20.3605} & \textbf{46.3988}\\
\midrule \fi
\textbf{LSTM}
& \eqwname & 0.1003 & 0.7066 & 0.1899 & 1.8919 & \textbf{-0.1264} & 0.1494 & 0.8386 & 7.5990 & 21.7948 & 49.8593\\
& \gfnamenb-Exp & \underline{0.1008} & \underline{0.7110} & \underline{0.1928} & \underline{1.9238} & \underline{-0.1450} & \underline{0.1370} & \underline{0.7369} & \underline{11.4428} & \underline{18.5193} & \underline{45.8684}\\
& \ppnamenb & \textbf{0.1049} & \textbf{0.7456} & \textbf{0.2122} & \textbf{2.0334} & -0.1745 & \textbf{0.1298} & \textbf{0.7290} & \textbf{12.4905} & \textbf{18.3295} & \textbf{45.3257}\\
\midrule
\textbf{GRU}
& \eqwname & 0.1122 & 0.9638 & 0.1841 & 1.6645 & -0.1740 & 0.1352 & 0.7090 & \underline{8.5684} & \underline{21.0297} & \underline{47.3572}\\
& \gfnamenb-Exp & \underline{0.1182} & \underline{1.0007} & \underline{0.1872} & \underline{1.7588} & \underline{-0.1207} & \underline{0.1281} & \underline{0.6688} & 8.4578 & 21.0399 & 47.4145\\
& \ppnamenb & \textbf{0.1183} & \textbf{1.0091} & \textbf{0.1928} & \textbf{1.7906} & \textbf{-0.1182} & \textbf{0.1250} & \textbf{0.6588} & \textbf{10.5918} & \textbf{20.1891} & \textbf{46.3092}\\
\ifodd \extraExps
\midrule
\textbf{ALSTM}
& \eqwname & 0.1091 & 0.8257 & 0.1954 & 1.6497 & \underline{-0.1399} & 0.1334 & 0.6965 & 8.6276 & \underline{20.1666} & 47.3265\\
& \gfnamenb-Exp & \underline{0.1100} & \underline{0.8360} & \underline{0.1985} & \underline{1.6684} & -0.1857 & \underline{0.1224} & \underline{0.6687} & \underline{9.5213} & 21.1583 & \underline{46.8636}\\
& \ppnamenb & \textbf{0.1106} & \textbf{0.8592} & \textbf{0.2005} & \textbf{1.7032} & \textbf{-0.1101} & \textbf{0.1217} & \textbf{0.6310} & \textbf{11.6311} & \textbf{18.4822} & \textbf{45.7709}\\
\fi
\bottomrule
\end{tabular}}
\caption{Performance comparison of the model-agnostic solutions on various scenarios and forecasting models.}
\label{tab:allresP1}
\end{threeparttable}
\end{table*}

\subsection{Experiments Results}

\subsubsection{Concept Drift Adaptation Methods Comparison (Q1)}
In this part, we compare \ppname with concept drift adaptation methods in different streaming data scenarios to answer {Q1} that \ppname is the SOTA in concept-drift-predictable scenarios. 
We compared all methods both in classification tasks (stock price trend forecasting) and regression tasks (electricity load forecasting, solar irradiance forecasting).

The following methods are compared:
\begin{itemize} 
    \item \textbf{\eqwnamenb}: Periodically \textbf{R}olling \textbf{R}etrain model on data in memory with equal weights. The memory only stores the most recent data in a limited window size.
    \item \textbf{\gfnamenb-Lin} \cite{koychev2000gradual}: Based on \eqwnamenb, \textbf{G}radual \textbf{F}orgetting by weights decaying \textbf{Lin}early by time.
    \item \textbf{\gfnamenb-Exp} \cite{klinkenberg2004learning}: Based on \eqwnamenb, \textbf{G}radual \textbf{F}orgetting by weights decaying \textbf{Exp}onentially by time.
    \item \textbf{ARF} \cite{gomes2017adaptive,gomes2018adaptive}: To deal with concept drift in the streaming data, \textbf{A}daptive \textbf{R}andom \textbf{F}orest does both internal and external concept drift detecting for each newly-created tree. The final prediction will be obtained by its voting strategy.
    \item \textbf{Condor} \cite{zhao2020handling}: \textbf{Con}cept \textbf{D}rift via m\textbf{o}del \textbf{R}euse is an ensemble method that handles non-stationary environments by both building new models and assigning weights for previous models.
    \item \textbf{\ppnamenb}: Our proposed method. \ppname is based on the setting of \eqwname and generates a new dataset by resampling to retrain forecasting models.
\end{itemize}

The compared methods can be categorized into two sets based on if they are model-agnostic. A model-agnostic solution can conduct concept drift adaptation without knowing the details of the forecasting model. Therefore, a model-agnostic solution can be applied to any model, which could be designed specifically to target scenarios. \eqwnamenb, \gfnamenb-Lin, \gfnamenb-Exp, \ppname are model-agnostic. They use the same forecasting model and historical data memory size. ARF and Condor are not model-agnostic. Condor is designed for classification, so it is not evaluated on the regression scenarios.

We use the open-source models released in Qlib's model zoo as our candidate forecasting model. On the validation data, LightGBM \cite{ke2017lightgbm} performs best on average. Therefore, we select LightGBM as our final forecasting model. 
The same forecasting model is also adopted by \eqwnamenb, \gfnamenb-Lin and \gfnamenb-Exp.
For the training process of \ppnamenb, we create tasks according to Figure \ref{fig:task}. The tasks are created in a more coarse-grained frequency to reduce the retraining cost. The time interval of two adjacent tasks is 20 trading days, 7 days and 7 days in stock price trend forecasting, electricity load forecasting and solar irradiance forecasting respectively. 
The test data still arrives in fine granularity.

The results are shown in Table \ref{tab:compareBaselines}. RR is the most naive method and simply periodically retrains models on data in memory. 
It performs worst among the compared methods.
\gfnamenb-Lin and \gfnamenb-Exp assume that the recent data distribution is more similar to the upcoming data. The weights of the most recent data are the highest and then decay by time. Such an assumption is valid in most datasets, and they outperform \eqwname in most scenarios.
ARF and Condor are the most recent SOTA solutions for concept drift adaptation. They adapt models to the most recent data to handle concept drift.
The forecasting performance is improved further.
However, the concept drift continues after adapting models to the most recent historical data. The adapted model could fail again when the concept drifts in the future. 
\ppname solves such a problem and performs best.
It models the trend of concept drifts 
and generates a new dataset whose distribution is closer to that in the future. Then the forecasting model is trained on the new dataset to handle the future concept drift. 

\subsubsection{Generalization to Different Forecasting Models (Q2)}

To answer {Q2}, we conduct experiments to demonstrate that \ppname can enhance the performance of different forecasting models in different streaming data scenarios.

The experiments involve different forecasting models. Therefore, we only compare model-agnostic concept drift adaptation solutions. For each forecasting model in each scenario, we compare \ppname with the \eqwname and GF-Exp (GF-Exp is an advanced version of GF-Lin. So GF-Lin is not included). 
We conduct experiments on multiple popular forecasting models, including \textbf{Linear} \cite{graybill1976theory},  \textbf{MLP} \cite{gardner1998artificial}, \textbf{LightGBM} \cite{ke2017lightgbm},
\ifodd \extraExps
\textbf{CatBoost} \cite{dorogush2018catboost}, 
\fi
\textbf{LSTM} \cite{hochreiter1997long}
\ifodd \extraExps
, \textbf{GRU} \cite{chung2014empirical} and \textbf{ALSTM} \cite{qin2017dual}. 
\else
and \textbf{GRU} \cite{chung2014empirical}. 
\fi

As the results shown in Table \ref{tab:allresP1}, \ppname outperforms others in most cases.  The results demonstrate that \ppname has captured the pattern of concept drift over time, which is the inherent nature of data and model-agnostic.

\subsubsection{Comparison of Optimization Methods}
\ppname converts the distribution distance optimization into a bi-level optimization by using a proxy model to model data distribution $p^{(t)}_{resam}(\bm{x}, y; \bm{\Theta})$. \ppname adapts a proxy model with a closed-form solution and convert Equation \eqref{eq:subject} into a differentiable operator. Therefore, the distribution distance between $D^{(t)}_{resam}(\bm{\Theta})$ and $D^{(t)}_{test}$ becomes differentiable.
Besides this approach, hyperparameter-optimization-based methods are eligible for the bi-level optimization. Gradient-based Hyperparameter Optimization (GHO) is one of them and adopted by a lot of research works \cite{maclaurin2015gradient,fan2020learning,baydin2017online}. Instead of requiring a proxy model with a closed-form solution, GHO only requires a  differentiable proxy model. It initializes the proxy model's parameters with $\phi_{(t),0}$ and then update it to $\phi_{(t),k}$ by $k$-steps with a gradient-based optimization method. At last, the $\phi_{(t),k}$ serves as $\phi_{(t)}$ in Equation \eqref{eq:subject} and makes Equation \eqref{eq:mse} differentiable. 

To compare different bi-level optimization algorithms, we create a new solution named \ppnamenb$^\dagger$ based on \ppname with GHO for optimization and an MLP proxy model. As the results show in Table \ref{tab:proxybase}, \ppnamenb$^\dagger$ outperforms \eqwnamenb, \gfnamenb-Lin and \gfnamenb-Exp. It demonstrates the effectiveness of our framework.
However, it slightly underperforms \ppnamenb.
\ppnamenb$^\dagger$ leverages a model with a larger capacity to model data distribution. It alleviates the problem of underfitting $p^{(t)}_{resam}(\bm{x}, y; \bm{\Theta})$. But the gap between the proxy model and the ground truth distribution still exists. The hyperparameter $k$ is sensitive.
A smaller $k$ may result in underfitting. A greater $k$ may waste computing resources and tend to overfit. 

\begin{table}
\centering
\resizebox{0.47\textwidth}{!}{%
\begin{tabular}{lrrrrr}
\toprule
\textbf{Method} & \textit{IC} & \textit{ICIR} & \textit{Ann.Ret.} & \textit{Sharpe} & \textit{MDD}\\
\midrule
\eqwname & 0.1092 & 0.8647 & 0.1803 & 1.5200 & -0.1797\\
\gfnamenb-Lin & 0.1090 & 0.8641 & 0.1880 & 1.5682 & \textbf{-0.1656}\\
\gfnamenb-Exp & 0.1091 & 0.8654 & 0.1889 & 1.6121 & -0.1738\\
\ppnamenb$^\dagger$ & \underline{0.1204} & \underline{0.9732} & \underline{0.2065} & \underline{1.9243} & \underline{-0.1705}\\
\ppnamenb & \textbf{0.1211} & \textbf{0.9921} & \textbf{0.2181} & \textbf{1.9409} & -0.1864\\
\bottomrule
\end{tabular}}
\caption{Comparison of \ppname with different bi-level optimization algorithms and proxy model; \ppnamenb$^\dagger$ is based on GHO and an MLP proxy model.}
\label{tab:proxybase}
\end{table}

\section{Conclusions and Future Works}
\label{sec:conclude}
In this paper, we propose a novel concept drift adaptation method \ppname to adapt models to future data distribution rather than latest historical data only like previous works.
\ppname models the concept drift and generates a new dataset by resampling historical data to guide the training process. Experiments on diverse scenarios demonstrate its effectiveness in concept-drift-predictable scenarios. \ifodd\hasappendix Moreover, we provided a detailed case study in Appendix \ref{Sec:FurtherExp} to show that the patterns learned by \ppname are reasonable and explainable.\fi \ppname is model-agnostic,so it can be transferred to any forecasting model used for downstream tasks.

Currently, \ppname is a static model that assumes the pattern of concept drift is static. This may not be valid in some scenarios (e.g. the concept drift frequency of $Task_{test}$ and $Task_{train}$ may be different).
For future work, we will improve \ppname to a dynamic one to solve this limitation.

\bibliography{aaai22}


\ifodd\hasappendix
\newpage

\appendix
\section{Notations}

To make it easier to follow the formulas in the paper, Table \ref{tab:notations} provides a notation list for Section \ref{sec:method_design}.

\begin{table*}[htbp]
\centering
\begin{tabular}{|c|l|}
\hline  
\textbf{Notation} & \hfill\textbf{Notation Explanation}\hfill{} \\
\hline
$\bm{X}$ & The feature of the streaming data. \\
\hline
$\bm{y}$ & The label of the streaming data. \\
\hline
$p_{train}^{(t)}\left(\bm{x}, y\right)$ / $p_{test}^{(t)}\left(\bm{x}, y\right)$ & The joint distribution of $\bm{x}$ and $y$ on the training/test data at the timestamp $t$. \\
\hline

$p_{train}^{(t)}\left(y \mid \bm{x}\right)$ / $p_{test}^{(t)}\left(y \mid \bm{x}\right)$ & The conditional distribution of $\bm{x}$ and $y$ on the training/test data at the timestamp $t$. \\
\hline

\multirow{2}{*}{$D^{(t)}_{train}$ / $D^{(t)}_{resam}$ / $D^{(t)}_{test}$} & The training/resampled/test dataset at timestamp $t$. \\
& $D^{(t)}_{resam}$ is resampled from $D^{(t)}_{train}$ with resampling probability $q^{(t)}_{train}$  \\
\hline
$task^{(t)}$ & The $t$-th task with training data $D^{(t)}_{train}$ and test data $D^{(t)}_{test}$. \\
\hline
$\mathcal{N}\left(\mu, \sigma\right)$ & The normal distribution with expectation $\mu$ and standard deviation $\sigma$. \\
\hline
$Task_{train}$ / $Task_{test}$ & A set of concept drift adaptation tasks. The tasks are generated on a rolling basis.\\
\hline
$\mathcal{M}_{\bm{\Theta}}$ & The \ppname model with parameters $\bm{\Theta}$. \\

\hline
\multirow{2}{*}{$y_{train}^{(t)}(\bm{x})$ / $y_{test}^{(t)}(\bm{x})$} & The expectation of $y$ under conditional distribution $p_{train}^{(t)}\left(y \mid \bm{x}\right)$ / $p_{test}^{(t)}\left(y \mid \bm{x}\right)$ \\
& on the training/test data of $task^{(t)}$. \\

\hline
$y_{resam}^{(t)}(\bm{x};\bm{\Theta})$ & The expectation of $y$ under the predicted conditional distribution of \ppname (i.e. $D_{resam}^{(t)}$). \\

\hline
\multirow{2}{*}{$q_{train}^{(t)}$} & The resampling probability of $task^{(t)}$, provided by \ppnamenb.\\
& It is used in the resampling process of $D_{train}^{(t)}$.\\
\hline
$L_{\bm{\Theta}}\left(task^{(t)}\right)$ & The loss of \ppname of $task^{(t)}$. \\

\hline
$\phi^{(t)}$ & The parameters of the proxy model for $task^{(t)}$. \\
\hline
\end{tabular}
\caption{Notations in Section \ref{sec:method_design}.}
\label{tab:notations}
\end{table*}

\section{Dataset Details}
\label{sec:dataDetails}
\subsection{Stock Trend forecasting}


The target of this task is to construct investment portfolios to maximize profit and minimize risk. An investment portfolio contains a basket of securities and cash. Portfolio management is constructing and balancing the investment portfolio periodically. Accurate forecasting of the securities' price trend is the key to successful portfolio management. Financial data is typically time-series.

The experiments are carried out on the stock data of the Chinese stock market, which includes time-series data of price from 2009 to 2020 in daily frequency. Besides the open, high, low, close price and trading volume data. About 300 fundamental and technical factors \cite{beyaz2018comparing} (a.k.a. features in Machine Learning) are included as well.
There are about 4 thousand different stocks.
We focus on investment in stocks in the experiments of the paper.
Due to the memory limitation in an online setting,
only two years of historical data are accessible in our experiment setting in our experiments. The test time range (evaluation time range) is from 2016 to 2020. 



\textbf{Evaluation and Metrics}

The metrics to evaluate a strategy are categorized into signal-based metrics and portfolio-based metrics.

The signal-based metrics are designed to evaluate the predictive abilities of a given forecasting. $IC$ (Information Coefficient) and $ICIR$ (IC Information Ratio) are the most typical metrics in this category. At each timestamp $t$, $IC$ could be measured by $IC^{(t)} = corr(rank(\hat{\bm{y}}^{(t)}), rank(\bm{ret}^{(t)}))$, where $\hat{\bm{y}}^{(t)}$ is the forecasting of stock price trend and $\bm{ret}^{(t)}$ is the ground truth of stock return.
For a given time period, a sequence of $IC$ could be measured $\bm{IC} = (IC^{(1)}, \dots, IC^{(T)})$. For that time period, $IC = mean(\bm{IC})$ and $ICIR = \frac {mean(\bm{IC})} {std(\bm{IC})}$. {\aamend The $rank$ function returns the ordered position of a given value within an array in ascending order. The $corr$ function is the correlation function, which is defined as:
\begin{equation*}
corr(\bm{x},\bm{y})=\frac{\sum_i (x_i-\bar{x})(y_i-\bar{y})}{\sqrt{\sum_i(x_i-\bar{x})^2\sum_i(y_i-\bar{y})^2}}
\end{equation*}
}

Besides signal-based metrics, portfolio-based metrics are widely used in practical investment analysis as well.
The portfolios are generated based on the forecasting of the stock price trends. Portfolio-based metrics evaluate the performance of portfolios by a specific trading strategy and backtest in the corresponding period. Portfolio-based metrics cover more factors than signal-based metrics, such as trading costs and portfolio weights. 
In our experiments, we use a  buying-winners-and-selling-losers investment strategy \cite{jegadeesh1993returns,wang2019alphastock}. The strategy and backtest system are based on an open-source quantitative investment platform Qlib \cite{yang2020qlib}. We will hold 50 stocks and trade daily. The benchmark index is CSI500 \cite{fan2017spillover}, and the metrics are based on the excess returns to the CSI500. Excess return measures how much a portfolio outperforms the benchmark.
The metrics include $Ann.Ret.$ (Annualized Return) \cite{briere2015virtual}, $Sharpe$ (Sharpe ratio) \cite{sharpe1994sharpe} and $MDD$ (Max Drawdown) \cite{chekhlov2004portfolio}, which are widely used in  quantitative investment to evaluate the performance of portfolios. $Ann.Ret.$  indicates the return of given portfolios each year. 
$Sharpe = \frac{Ann.Ret.}{Ann.Vol.}$ and  ${Ann.Vol.}$ indicates the annualized volatility.
To achieve higher $Sharpe$, portfolios are expected to maximize the total excess return and minimize the volatility of the daily excess returns.
$MDD$ is the maximum relative asset value loss to the previous peak, and the lower $abs(MDD)$ is preferred.
The trading strategies will hold the 50 stocks with the highest forecasting scores on the first trading day in the backtesting process. For each following trading day, stocks with the lowest forecasting score inside the current portfolio will be replaced by the stock with the highest forecasting score outside. The transaction cost(slippage is included) is $2$ \textperthousand.

\subsection{Electricity Load Forecasting}
Electricity load forecasting plays an essential role in the electricity corporation arrangement. Operation and maintenance costs can be greatly reduced by accurate forecasting, which improves the reliability of the power supply and the power transmission systems.

The electricity load dataset in our experiments is a UCI repository dataset that contains electricity usage from 370 users between 2011 and 2015 with a quarter-hour frequency. The memory limitation is one year of data for electricity load forecasting. In our experiments, the evaluation time range is from 2013 to 2015.

\textbf{Evaluation and Metrics}

The metrics used in electricity load forecasting are $NMAE$ (Normalized Mean Absolute Error) and $NRMSE$ (Normalized Root Mean Square Error), which are the same with \cite{yu2016temporal,salinas2020deepar}:
\begin{equation*}
    NMAE = \frac{\sum_{i=1}^{N} \left|{y}_{i}-f\left(\bm{x}_i\right)\right|}{\sum_{i=1}^{N}  \left|{y}_{i}\right|}
\end{equation*}
\begin{equation*}
    NRMSE = \frac{\sqrt{\frac{1}{N}\sum_{i=1}^{N} {\left({y}_{i}-f\left(\bm{x}_i\right)\right)}^2}}{\frac{1}{N} \sum_{i=1}^{N} \left|{y}_{i}\right|}
\end{equation*}
where $N$ is the number of samples, and the $i$-th sample is represented as $\left(\bm{x}_i, y_i\right)$.


\subsection{Solar Irradiance Forecasting}

Solar irradiance forecasting reduces solar power generation's uncertainty caused by the intermittence of solar irradiance. Solar irradiance forecasting can be used to schedule power generation and balance energy production and consumption.

The solar irradiance forecasting dataset \cite{pedro2019comprehensive} is collected from 2014 to 2016 with a 5-minute frequency. It combines both irradiance features and sky-image features. The memory limitation is one year of data. In our experiments, the time range of test data is from 2015 to 2016.

\textbf{Evaluation and Metrics}

The metrics include $Skill$ (forecast skill), $MAE$ (Mean Absolute Error) and $RMSE$ (Root Mean Square Error), which are specified in \cite{pedro2019comprehensive}:
\begin{equation*}
    MAE = \frac{1}{N}\sum_{i=1}^{N} \left|{y}_i-f\left(\bm{x}_i\right)\right|
\end{equation*}
\begin{equation*}
    RMSE = \sqrt{\frac{1}{N}\sum_{i=1}^{N} {\left({y}_{i}-f\left(\bm{x}_i\right)\right)}^2}
\end{equation*}
\begin{equation*}
    skill \simeq 1 - \frac{RMSE}{RMSE_p}
\end{equation*}
where $RMSE_p$ is the $RMSE$ of the persistence forecasting results, which is already provided in the dataset.


\section{Model Details}
\label{sec:ModelDetails}

In this section, we will introduce some model details of \ppname for better reproducibility.

\subsection{Detailed Feature Design}
We briefly described the feature design of \ppname in Section \ref{sec:feature_design}. In this section, we will introduce the detailed feature design.
Because \ppname tries to capture the pattern of concept drift over time, the change of data distribution over time is the critical information for $\mathcal{M}_{\Theta}$.
Feature extractor $g$ is rule-based currently and extracts features based on the similarity of data in different periods. $g$ output a matrix with shape $\langle sample, feature\rangle$. It will extract features for each sample.  The extracted features of each sample $\bm{x}$ can be represented as $g_{\bm{x}}\left(D^{(t)}_{train}\right) = \left(sim^{(\tau)}, sim^{(\tau - 1)}, \dots, sim^{(\tau - k)}\right)$, where $\tau$ represents the timestamp closest to previous test data in $D^{(t)}_{train}$ (now it has become training data at $t$) and 
$sim^{(\cdot)}$ 
represents the similarity of data
$\bm{x}$
to the data $D^{(\tau)}_{test}$. {\amend Kullback–Leibler divergence is a widely-used metric for distribution distance, which is adopted as the similarity metric, as well as the optimization target in our paper.} The sequence of data similarity reflects the change in distribution of streaming data.
For more robustness in statistics, we use all the data in the short period where the sample $\bm{x}$ belongs to calculate the data similarity.

\section{Further Experimental Results}
\label{Sec:FurtherExp}

The experiments are carried out on the machines with Intel Xeon Platinum 8171M Processors and Tesla K80 GPU.

\subsection{Experiment Running Time}
\begin{table}[H]
\centering
\begin{threeparttable}[b]
\begin{tabular}{l|r|r|r}
\toprule
{Method} & {Stock Trend} & {Electricity} & {Solar}\\
\midrule
\eqwname & 185.37 & 160.40 & 146.38 \\
\gfnamenb-Lin & 194.63 & 188.49 & 166.90\\
\gfnamenb-Exp & 196.65 & 174.63 & 147.50\\
ARF & 1377.53 & 760.20 & 570.00\\
Condor & 34.47 & --- & ---\\
\ppnamenb & 216.50 & 185.53 & 158.23\\
\bottomrule
\end{tabular}
\caption{Running time (minutes) comparison of the concept drift adaptation methods. ARF is slower than other methods because its toolkit does not support parallelism. Condor is an ensemble method that doesn't retrain models, so its time cost is lower than other retraining methods.}
\label{tab:timeBaselines}
\end{threeparttable}
\end{table}

\subsection{Performance on the Unpredictable Concept Drift}
In addition to the concept-drift-predictable real-world scenarios, we conduct experiments on concept-drift-unpredictable scenarios in this part. We construct 4 regression streaming datasets with random/unpredictable abrupt concept drifts.
{\aamend For each period under a certain concept, features $\bm{x}\in\mathbb{R}^{N\times d}$ are randomly generated, where $x_{i,j}\sim\mathcal{N}(0,I)$. The corresponding true labels are $\bm{y}=\bm{x}\bm{W}+\epsilon$, where $\bm{W}\in\mathbb{R}^d$ is randomly generated, $\|\bm{W}\|^2_2=1$ and $\epsilon$ is a random noise from standard normal distribution.} We changes the regression model randomly to simulate random abrupt concept drift. 
The experiment results in Figure \ref{fig:abruptCD}. The horizontal axis represents different generated datasets and the average results. The vertical axis represents the correlation between the adapted model's forecasting and the labels (the higher the better).
It shows that \ppname cannot achieve stable and good performance compared to others. \ppname is designed for predictable concept drift and fails to handle unpredictable concept drift which is out of our scope. 

\begin{figure}[ht]
\centering
  \includegraphics[width=0.8\linewidth]{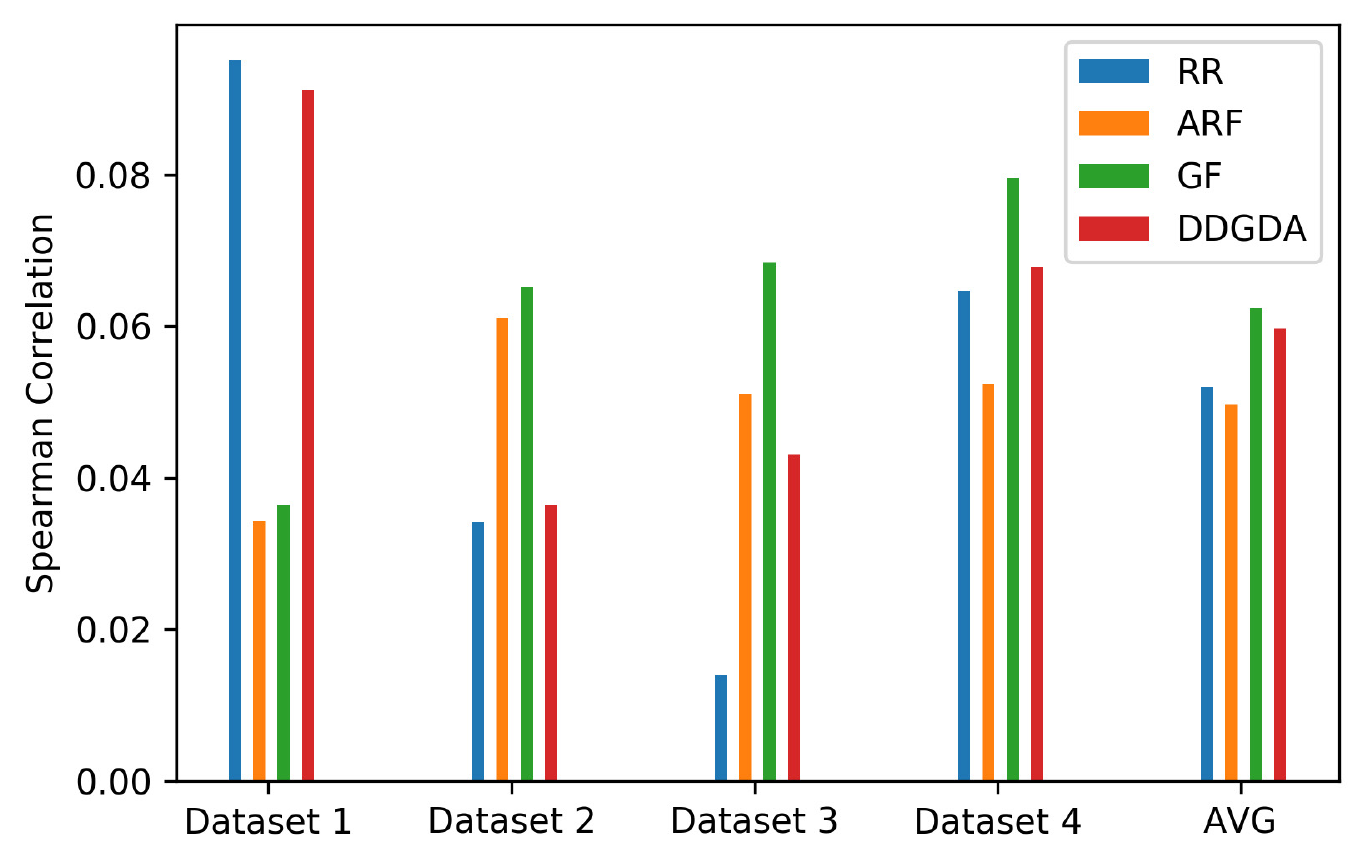}
  \caption{Comparison of different methods in 4 concept-drift-unpredictable scenarios (random abrupt concept drift). \ppname can't achieve stable and good performance compared to others.}
  \label{fig:abruptCD}
\end{figure}

{\aamend
\subsubsection{Effects of Varying Time Intervals}

The best rolling time interval depends on the frequency of the concept drift, which is an inherent nature of the dataset. In Figure \ref{fig:timeinterval}, we conduct experiments to show the relationship between different time intervals and final performance in the stock trend forecasting scenario. It shows that smaller interval is helpful to the performance. But the marginal improvement is much smaller when the interval is small. Users can create a validation set to get this curve and pick the elbow.

\begin{figure}[H]
\centering
  \includegraphics[width=\linewidth]{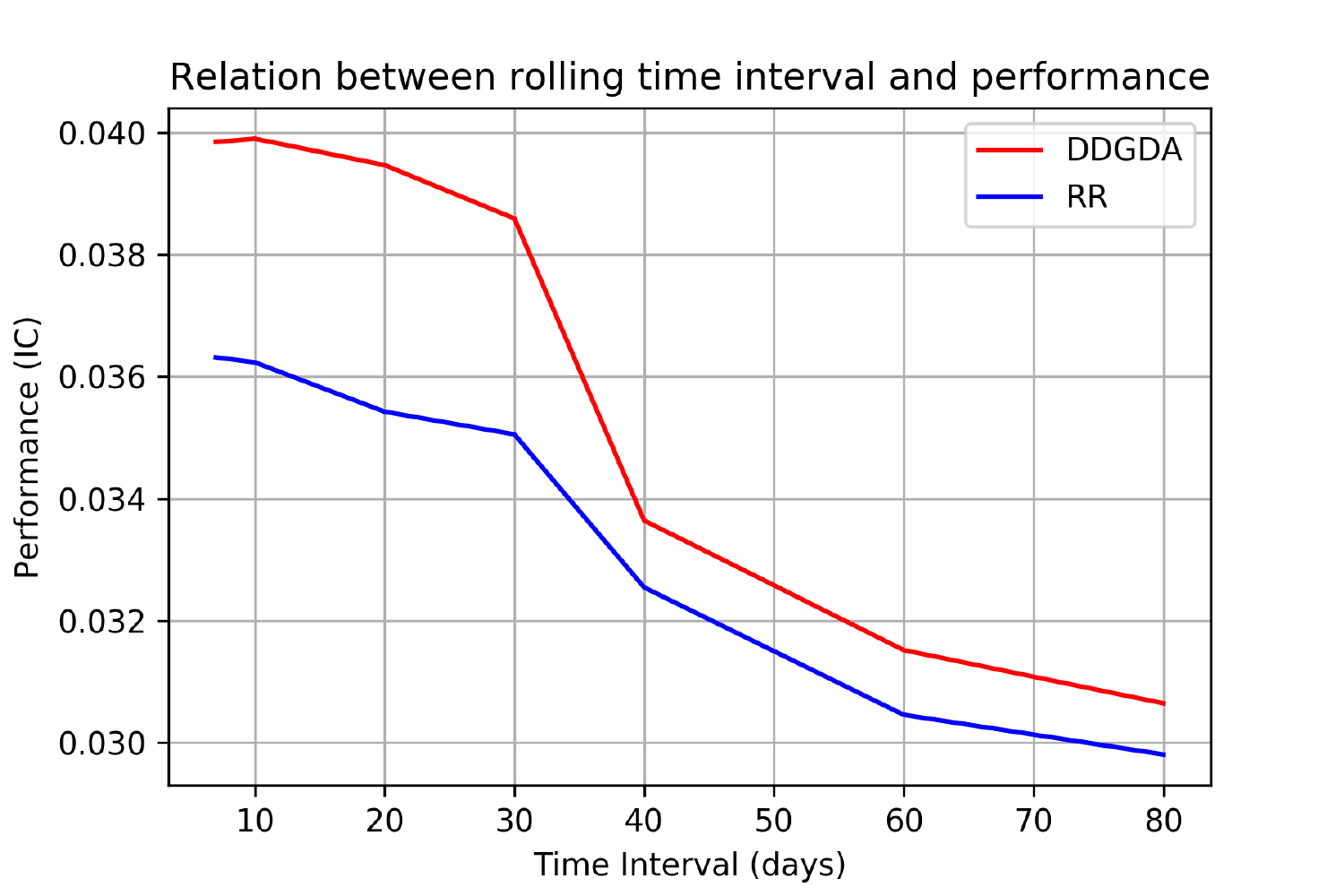}
  \caption{The relation between different time intervals and performance on stock price trend forecasting.}
  \label{fig:timeinterval}
\end{figure}

}
\subsection{Case Study}
\label{Sec:casestudy}

In this section, we will conduct some case studies and explain the learned pattern of \ppnamenb. The financial data are highly dynamic and volatile, making it an excellent scenario to present our case studies.

\begin{figure}[htbp]
  \includegraphics[width=\linewidth]{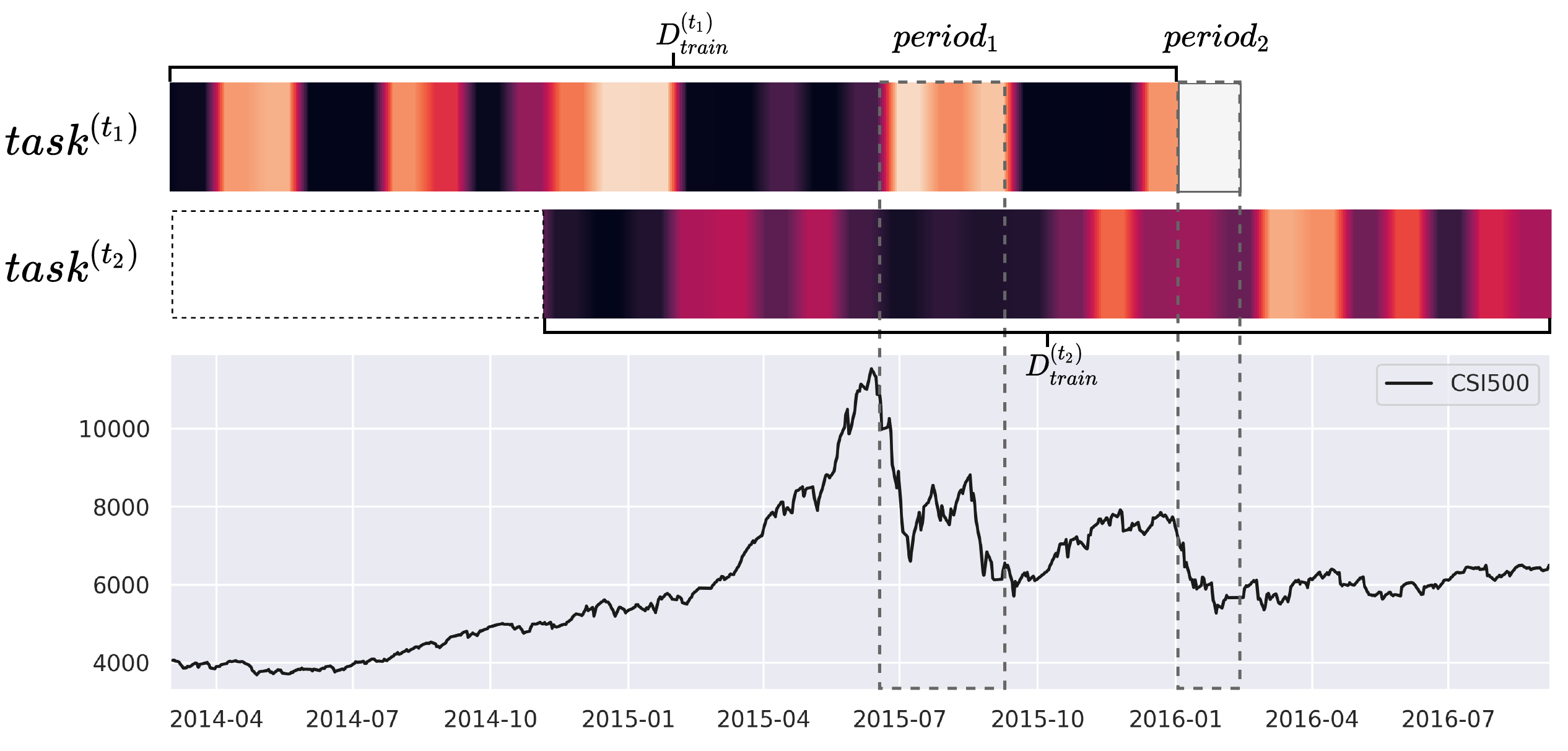}
  \caption{The data resampling probability is given by \ppname on the stock forecasting task in different tasks. \ppname gives different resampling probability to the time-series data, the data resampling probability distribution changes dynamically according to stock market states at different timestamps. \ppname gives more attention to the data with a similar descending price trend. 
  }
  \label{fig:casestudy}
\end{figure}

Figure \ref{fig:casestudy} demonstrates several case studies of \ppnamenb.
The two heatmaps on the top indicate two tasks like Figure \ref{fig:task}. 
The brightness of the color indicates the distribution of the data resampling probability given by \ppnamenb. The brighter the color, the higher the probability. The data resampling probability changes over time.
The figure below is the price trend of the CSI500\footnote{http://www.csindex.com.cn/en/indices/index-detail/000905}, which is a stock index to represents the trend of the stock market.
The $x$-axis of all the figures is aligned, which can help us analyze the behaviors of \ppname at different times.  

As we can see on $task^{(t_2)}$, \ppname gives a higher probability to the recent data. Such behavior is similar to \gfnamenb-Lin and \gfnamenb-Exp.
The probability changes non-monotonically because \ppname leverages more information in data and captures more intricate patterns. 
The color during $period_1$ is dark, which indicates a low resample probability on such data. The index trend below suggests that the stock market is suffering a great depression at $period_1$. The data distribution in such a particular market state is quite different, and \ppname gives quite a reasonable resample probability.

As a concept drift adaptation solution, \ppname can adapt the forecasting model to the dynamic data distribution. The state of the market changes dynamically. Therefore, the optimal resample probability reweighting distribution changes over time. 
$task^{(t_1)}$ demonstrates another case of resample probability given by \ppnamenb.
The latest data in $D_{train}^{(t_1)}$ is nearly at the very beginning of the stock market crash in early 2016.
Thanks to the forward-looking design, \ppname is aware of concept drift and gives more attention to the data with a similar pattern (i.e., the descending price trend). For data in $period_1$, \ppname gives a total different resample probability in $task^{(t_1)}$ than that in $task^{(t_2)}$. 

\begin{figure}[htbp]
  \includegraphics[width=\linewidth]{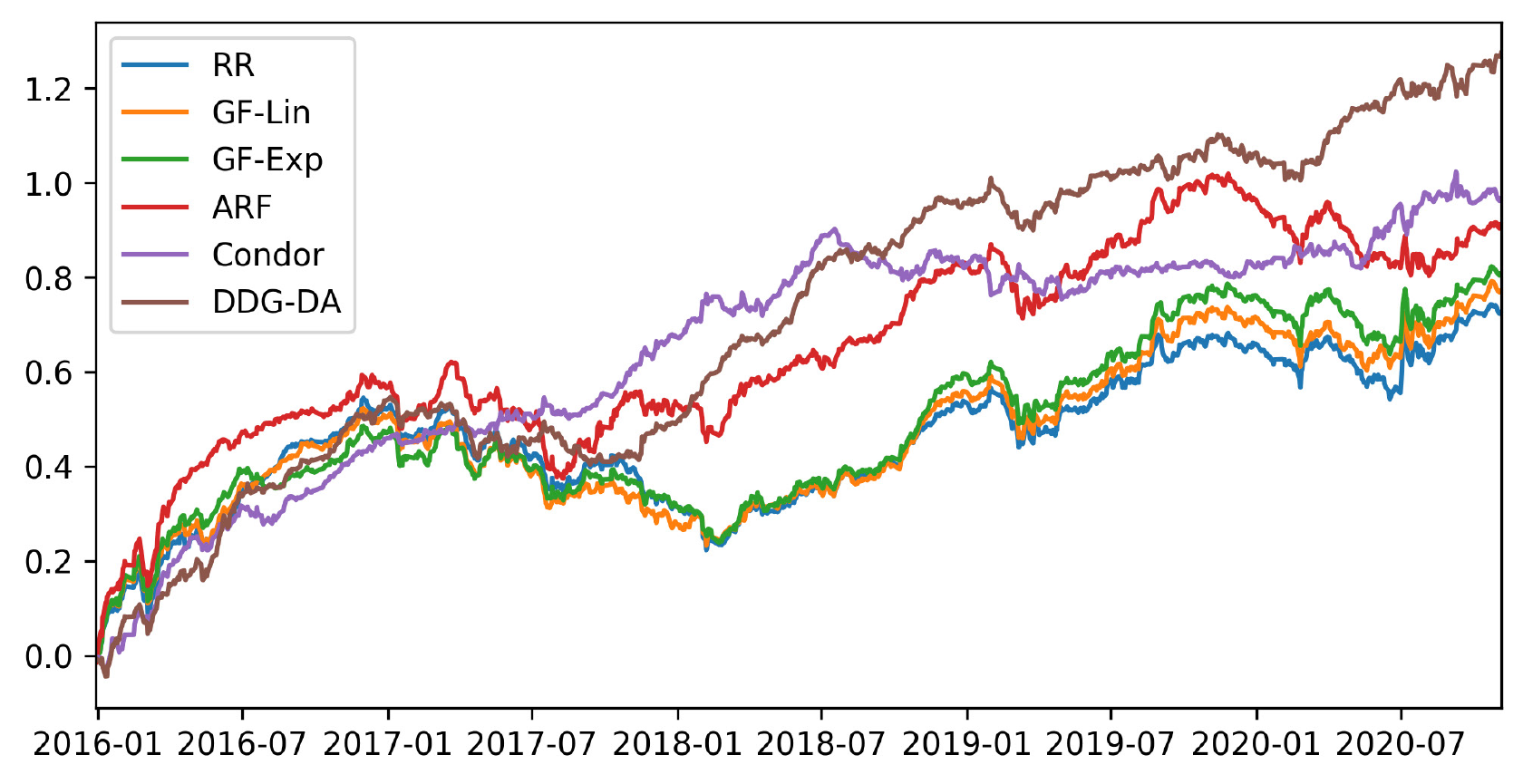}
  \caption{The accumulated excess return curves (the benchmark is CSI500) in stock price trend forecasting of different concept drift adaptation solutions from Table \ref{tab:compareBaselines}. It shows that \ppname performs better than other methods. 
  }
  \label{fig:returncurve}
\end{figure}

Figure \ref{fig:returncurve} demonstrates the accumulated return curves of different concept drift  adaptation solutions from Table \ref{tab:compareBaselines}. 
The benchmark is CSI500.
To avoid the curve skewed exponentially by current assets value, the accumulated return is the summation of daily returns instead of a cumulative product.
\eqwnamenb, \gfnamenb-Lin and \gfnamenb-Exp adopt a straightforward concept drift adaptation solution. Much detailed information is lost, and the value of their portfolios suffered a significant drawdown in 2017.
Other more sophisticated concept drift adaptation methods avoid such a drawdown and achieve better performance.
\ppname gains the highest returns among them. 
Besides achieving higher profit, the $MDD$ and $Sharpe$ are better. It indicates that the volatility of the portfolio's returns is low, which results in lower risk. Such volatility is the result of the model's failure to adopt the new data distribution.
These results demonstrate that \ppname has superior and stable performance on predictable concept drift adaptation.


\section{Algorithm Analysis}
\label{Sec:theoretical}
\begin{theorem}
Given two normal distributions $q\left(z\right)\sim\mathcal{N}\left(\mu_1,\sigma_1^2\right)$, $p\left(z\right)\sim\mathcal{N}\left(\mu_2,\sigma_2^2\right)$, $D_{KL} \left( q\left(z\right) \parallel  p\left(z\right) \right)$ is proportional to the mean square error of $q\left(z\right)$ and $p\left(z\right)$ when ${\sigma}_1$ and ${\sigma}_2$ are constants. 
\label{theo:01}
\end{theorem}

\begin{equation}
\begin{aligned}
& D_{K L}\left(q\left(z\right) \| p\left(z\right)\right) \\
=&\int q\left(z\right) \log \frac{q\left(z\right)}{p\left(z\right)} \mathrm{d} z \\
=&\int q\left(z\right)\left[\log q\left(z\right)-\log p\left(z\right)\right] \mathrm{d} z \\
=&\underbrace{\int q\left(z\right) \log q\left(z\right) \mathrm{d} z}_{\text{Term 1}}-\underbrace{\int q\left(z\right) \log p\left(z\right) \mathrm{d} z}_{\text{Term 2}} \\
=&\underbrace{\left(-\frac{1}{2} \log (2 \pi)-\frac{1}{2} \left(\log \sigma_{1}^{2}+1\right)\right)}_\text{Equation \eqref{eq:term1}}\\
&-\underbrace{\left(-\frac{1}{2} \log (2 \pi)-\frac{1}{2} \log \sigma_{2}^{2}-\frac{1}{2} \left[\frac{\sigma_{1}^{2}}{\sigma_{2}^{2}}+\frac{\left(\mu_{1}- \mu_{2}\right)^{2}}{\sigma_{2}^{2}}\right]\right)}_\text{Equation \eqref{eq:term2}}\\
\propto&{\left(\mu_1-\mu_2\right)}^2
\end{aligned}
\label{eq:theorem_proof}
\end{equation}
\begin{equation}
\begin{aligned}
&\underbrace{\int q(z) \log q(z) \mathrm{d} z}_{\text{Term 1}} \\
=&\int \mathcal{N}\left(z ; \mu_{1}, \sigma_{1}^{2}\right) \log \mathcal{N}\left(z ; \mu_{1}, \sigma_{1}^{2}\right) \mathrm{d} z \\
=& E_{q\left(z\right)}\left[\log \frac{1}{\sqrt{2 \pi \sigma_{1}^{2}}} \exp \left(-\frac{\left(z-\mu_{1}\right)^{2}}{2 \sigma_{1}^{2}}\right)\right] \\
=& E_{q\left(z\right)}\left[-\frac{1}{2} \log \left(2 \pi \sigma_{1}^{2}\right)-\frac{\left(z-\mu_{1}\right)^{2}}{2 \sigma_{1}^{2}}\right] \\
=&-\frac{1}{2} \log \left(2 \pi \sigma_{1}^{2}\right)-E_{q\left(z\right)}\left[\frac{\left(z-\mu_{1}\right)^{2}}{2 \sigma_{1}^{2}}\right] \\
=&-\frac{1}{2} \log (2 \pi)-\frac{1}{2} \log \sigma_{1}^{2}-\frac{1}{2}  E_{q\left(z\right)}\left[\frac{\left(z-\mu_{1}\right)^{2}}{\sigma_{1}^{2}}\right] \\
=&-\frac{1}{2} \log (2 \pi)-\frac{1}{2} \log \sigma_{1}^{2}-\frac{1}{2\sigma_{1}^{2}} E_{q\left(z\right)}\left[\left(z-\mu_{1}\right)^{2}\right]\\
=&-\frac{1}{2} \log (2 \pi)-\frac{1}{2} \log \sigma_{1}^{2}-\frac{1}{2\sigma_{1}^{2}} \sigma_{1}^{2}\\
=&-\frac{1}{2} \log (2 \pi)-\frac{1}{2} \left(\log \sigma_{1}^{2}+1\right)
\end{aligned}
\label{eq:term1}
\end{equation}
\begin{equation}
\resizebox{\linewidth}{!}{$
\begin{aligned}
&\underbrace{\int q(z) \log p(z) \mathrm{d} z}_{\text{Term 2}} \\
=&\int \mathcal{N}\left(z ; \mu_{1}, \sigma_{1}^{2}\right) \log \mathcal{N}\left(z ; \mu_{2}, \sigma_{2}^{2}\right) \mathrm{d} z \\
=&E_{q\left(z\right)}\left[\log \frac{1}{\sqrt{2 \pi \sigma_{2}^{2}}} \exp \left(-\frac{\left(z-\mu_{2}\right)^{2}}{2 \sigma_{2}^{2}}\right)\right] \\
=&E_{q\left(z\right)}\left[-\frac{1}{2} \log \left(2 \pi \sigma_{2}^{2}\right)-\frac{\left(z-\mu_{2}\right)^{2}}{2 \sigma_{2}^{2}}\right] \\
=&-\frac{1}{2} \log \left(2 \pi \sigma_{2}^{2}\right)-E_{q\left(z\right)}\left[\frac{\left(z-\mu_{2}\right)^{2}}{2 \sigma_{2}^{2}}\right] \\
=&-\frac{1}{2} \log (2 \pi)-\frac{1}{2} \log \sigma_{2}^{2}-\frac{1}{2} E_{q\left(z\right)}\left[\frac{\left(z-\mu_{2}\right)^{2}}{\sigma_{2}^{2}}\right] \\
=&-\frac{1}{2} \log (2 \pi)-\frac{1}{2} \log \sigma_{2}^{2}-\frac{1}{2\sigma_{2}^{2}} E_{q\left(z\right)}\left[\left(z-\mu_{2}\right)^{2}\right]\\
=&-\frac{1}{2} \log (2 \pi)-\frac{1}{2} \log \sigma_{2}^{2}-\frac{1}{2\sigma_{2}^{2}} E_{q\left(z\right)}\left[z^{2}-2 z \mu_{2}+\mu_{2}^{2}\right]\\
=&-\frac{1}{2} \log (2 \pi)-\frac{1}{2} \log \sigma_{2}^{2}- \frac{1}{2\sigma_{2}^{2}}\left[E_{q\left(z\right)} z_{j}^{2}-2 E_{q\left(z\right)} z \mu_{2}+E_{q\left(z\right)} \mu_{2}^{2}\right]\\
=&-\frac{1}{2} \log (2 \pi)-\frac{1}{2} \log \sigma_{2}^{2}- \frac{1}{2\sigma_{2}^{2}}\left[\sigma_{1}^{2}+\mu_{1}^{2}-2 \mu_{1} \mu_{2}+\mu_{2}^{2}\right]\\
=&-\frac{1}{2} \log (2 \pi)-\frac{1}{2} \log \sigma_{2}^{2}- \frac{1}{2\sigma_{2}^{2}}\left[\sigma_{1}^{2}+\left(\mu_{1}-\mu_{2}\right)^{2}\right]\\
=&-\frac{1}{2} \log (2 \pi)-\frac{1}{2} \log \sigma_{2}^{2}-\frac{1}{2} \left[\frac{\sigma_{1}^{2}}{\sigma_{2}^{2}}+\frac{\left(\mu_{1}-\mu_{2}\right)^{2}}{\sigma_{2}^{2}}\right]
\end{aligned}
$}
\label{eq:term2}
\end{equation}

\fi

\end{document}